\documentclass[runningheads]{llncs}

\usepackage{eccvabbrv}

\usepackage{graphicx}
\usepackage{booktabs}
\usepackage{xspace}
\usepackage[dvipsnames]{xcolor}
\usepackage[accsupp]{axessibility}  
\usepackage{amssymb}
\usepackage{pifont}

\definecolor{citecolor}{HTML}{0071bc}
\definecolor{linkcolor}{RGB}{215,   0,   64}
\usepackage[pagebackref,breaklinks=true,letterpaper=true,colorlinks=true,linkcolor=linkcolor,citecolor=citecolor,bookmarks=false]{hyperref}

\usepackage{graphicx}
\usepackage{amsmath}
\usepackage{url}            
\usepackage{amsfonts}       
\usepackage{nicefrac}       
\usepackage{microtype}      
\usepackage{algorithm}
\usepackage{comment}
\usepackage{pifont}
\usepackage[noend]{algpseudocode}
\usepackage{bbm}
\usepackage{dsfont}
\usepackage{easylist}
\usepackage{xspace}
\usepackage{wrapfig}
\usepackage{multirow} 
\usepackage{amssymb}
\usepackage{caption}
\usepackage{subcaption}
\usepackage{float}
\usepackage{array}
\usepackage{hyperref}
\usepackage{orcidlink}

\newcommand{\mmfm}{MLLM\xspace}
\newcommand{\ttc}{Talk2Car\xspace}
\newcommand{\dLM}{DriveLM\xspace}
\newcommand{\dataset}{LV3D\xspace}
 
\definecolor{skyblue1}{RGB}{114, 159, 207}
\definecolor{myred}{RGB}{205, 30, 30}
\definecolor{mygreen}{RGB}{20,   132,   0}
\definecolor{mytangoblue}{RGB}{59, 109, 172}
\definecolor{mytangoorange}{RGB}{243, 121, 33}
\definecolor{mypurple}{RGB}{150,   20,   150}
\definecolor{mygrey}{RGB}{150,   150,   150}
\definecolor{mycell1}{RGB}{240,   230,   230}
\definecolor{mycell2}{RGB}{240,   230,   230}
\definecolor{darkgrey}{RGB}{80,   80,   80}
\definecolor{specialist_green}{RGB}{113, 129, 112}
\definecolor{my_yellow}{RGB}{229, 193, 65}

\newcommand{\myparagraph}[1]{\vspace{1pt}\noindent{\bf #1}}
\newcommand{\xmark}{\ding{55}}

\newlength\savewidth

\newcommand{\ours}{\textbf{\scshape{Cube-LLM}}\xspace}

\begin{document}

\title{Language-Image Models with 3D Understanding} 

\titlerunning{\ours}

\author{Jang Hyun Cho\inst{1,2}\and
Boris Ivanovic\inst{2}\and
Yulong Cao\inst{2} \and
Edward Schmerling\inst{2} \and\\
Yue Wang\inst{2} \and 
Xinshuo Weng\inst{2} \and
Boyi Li\inst{2}\and
Yurong You\inst{2} \and\\
Philipp Krähenbühl\inst{1,\star} \and
Yan Wang\inst{2,\star} \and
Marco Pavone$^{2,}$\thanks{Equal advising}
}

\authorrunning{JH Cho et al.}

\institute{UT Austin \and NVIDIA Research \\
\email{janghyuncho7@utexas.edu}\\
\email{\{bivanovic,yulongc,eschmerling,yuewang,xweng,\\
boyil,yurongy,yanwan,mpavone\}@nvidia.com}\\
}
\maketitle

\begin{abstract}
Multi-modal large language models (MLLMs) have shown incredible capabilities in a variety of 2D vision and language tasks.
We extend \mmfm{}s' perceptual capabilities to ground and reason about images in 3-dimensional space. 
To that end, we first develop a large-scale pretraining dataset for 2D and 3D called \dataset{} by combining multiple existing 2D and 3D recognition datasets under a common task formulation: as multi-turn question-answering.
Next, we introduce a new \mmfm{} named \ours and pre-train it on \dataset{}. 
We show that pure data scaling makes a strong 3D perception capability without 3D specific architectural design or training objective. 
\ours exhibits intriguing properties similar to LLMs: (1) \ours can apply chain-of-thought prompting to improve 3D understanding from 2D context information.
(2) \ours can follow complex and diverse instructions and adapt to versatile input and output formats. 
(3) \ours can be visually prompted such as 2D box or a set of candidate 3D boxes from specialists.
Our experiments on outdoor benchmarks demonstrate that \ours significantly outperforms existing baselines by 21.3 points of AP$_{\text{BEV}}$ on the \ttc dataset for 3D grounded reasoning and 17.7 points 
on the \dLM dataset for complex reasoning about driving scenarios, respectively. 
\ours also shows competitive results in general MLLM benchmarks such as refCOCO for 2D grounding with (87.0) average score, as well as visual question answering benchmarks such as VQAv2, GQA, SQA, POPE, etc. for complex reasoning.  
Our project is available at \url{https://janghyuncho.github.io/Cube-LLM}.

\end{abstract}

\section{Introduction}
\label{sec:intro}

Internet-scale visual data have brought forth the advent of multi-modal large language models (MLLMs). 
Rich and diverse visual supervision aligns pre-trained large language models with billions of parameters to visual modality. 
The best MLLMs can recognize, understand, and reason about images and videos far better than any of specially designed architectures and algorithms~\cite{gpt4_V,team2023gemini}. 
The decades worth of computer vision datasets  ---image classification, captioning, object detection, grounding, document parsing, optical character recognition (OCR)--- fuels the powerful MLLMs through jointly training as a {next token prediction} task. 
Introducing the ability to ``ground'' in 2-dimensional space (\emph{image coordinates}) bridges the low-level perception to high-level reasoning about visual input, much like human cognition. 
However, one critical difference is that we perceive the world in 3-dimensional space (\emph{view coordinates}). 
This 3-dimensional grounding allows us to perceive and reason about the visual input closer to the actual world, which the current state of MLLMs has not explored yet. 
\begin{figure}[!t]
    \centering
    \includegraphics[width=\linewidth]{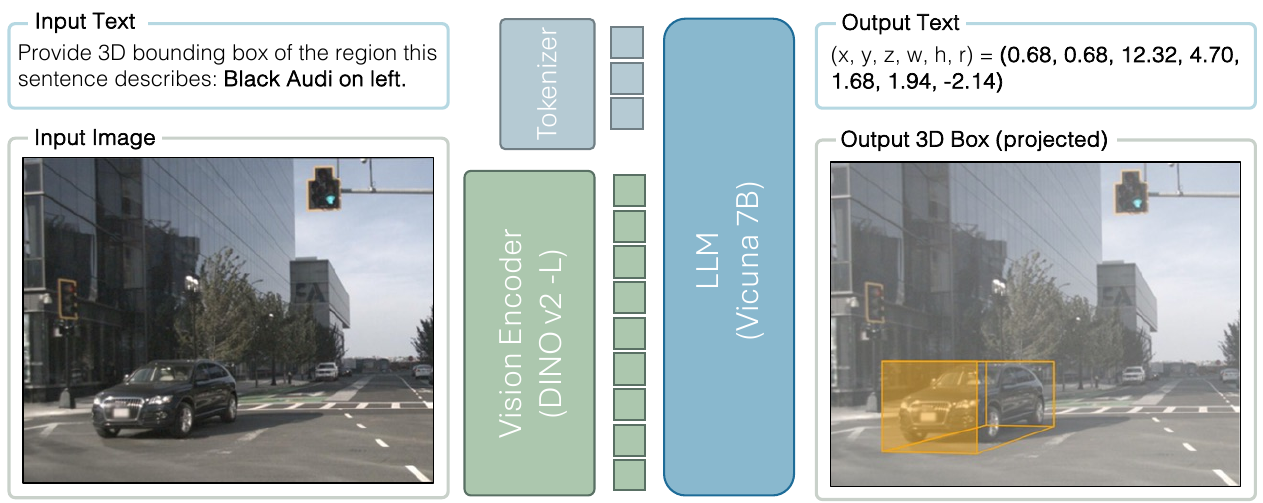}
    \caption{
    The overview of \ours for 3D-grounded reasoning. The task requires a model to take an image, understand the input text prompt (e.g., ``\textit{Black Audi on left.''}) and ground it in 3-dimensional space. 
    }
    \label{fig:teaser-figure}
    \vspace{-0.5cm}
\end{figure}

\begin{figure}[!h]
    \centering
    \includegraphics[width=\linewidth]{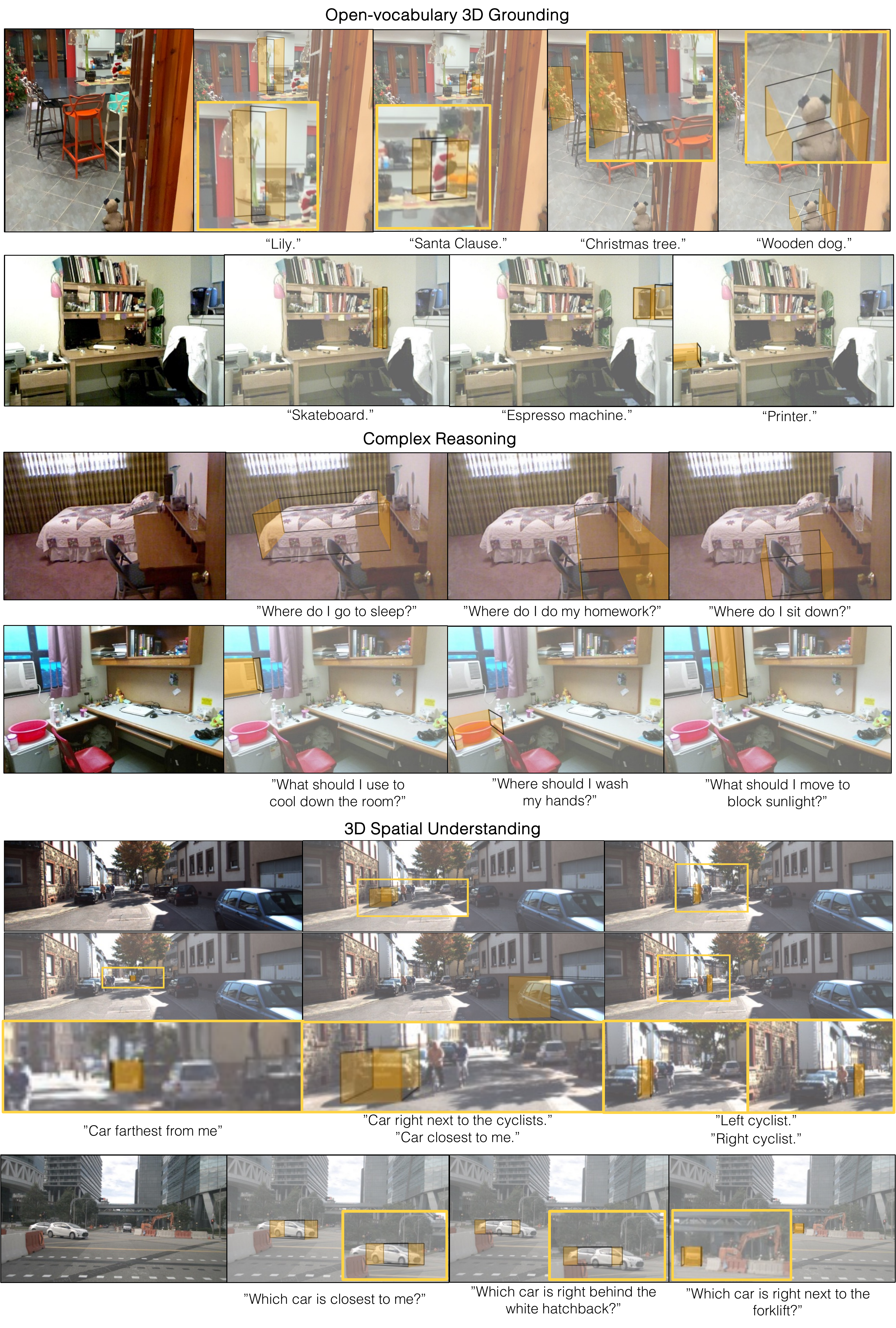}
     \vspace{-0.5cm}
    \caption{\textbf{Qualitative results} of \ours 3D grounding in 3 aspects: \emph{open-vocabulary understanding} (\textbf{top}), \emph{complex reasoning} (\textbf{middle}), and \emph{3D spatial understanding} (\textbf{bottom}). Best viewed in \textcolor{my_yellow}{\textbf{color}}, \textbf{zoomed}. }
    \label{fig:vis}
    \vspace{-1.1cm}
\end{figure}

In this work, our goal is to develop a framework to train a MLLM capable of reasoning in both 2D and 3D spaces. 
We demonstrate that pure data scaling can achieve our goal without any 3D specific architectural design or training objective. 
We instead focus on careful data curation to address one question: \emph{what tasks will induce 2D to 3D generalization?} 
To this end, we introduce a large scale language-image pretraining dataset for 2D and 3D, called LV3D. 
We start with combining a diverse collection of 2D and 3D vision datasets for indoors and outdoors and standardize labels to follow the consistent format across datasets.
We blend in the vision datasets with instruction-following data of MLLM training as a series of question-answer pairs ($\S$~\ref{sec:data_scale}). 
Next, we augment our blended datasets by {decomposing} the vision labels into easier tasks (\emph{e.g., 3D box $\rightarrow$ 2D point, depth, size, orientation}).
This trains our model to adapt to versatile input and output formats, and connects the underlying 2D and 3D structure ($\S$~\ref{sec:task_scale}). 
Most importantly, we mix in a series of QA pairs about an object for ``step-by-step'' reasoning, from easier (\emph{e.g., 2D box}) to harder (\emph{e.g., 3D box}) task.
This directly induces 2D to 3D generalization due to autoregressive nature of MLLMs ($\S$~\ref{sec:v_cot}).
Finally, we train a MLLM on LV3D as a single ``next token prediction'' task, called \ours ($\S$~\ref{sec:arch}). 

\ours exhibits a number of intriguing properties.
First, \ours can self-improve its 3D reasoning performance by prompting with its own 2D predictions.
This \emph{visual chain-of-thought reasoning} resembles the well-known behavior of LLMs~\cite{wei2022chain}. 
Second, \ours can adapt to versatile input and output formats and questions, which follows \emph{instruction following} ability of LLMs~\cite{wei2022finetuned}. 
Finally, \ours can be \emph{prompted} with any specialist models for any additional modalities (\emph{e.g., LiDAR}) by simply adding their predictions to the question. 
\ours shows remarkable improvement with data-scaling in both 2D and 3D, for indoor and outdoor scene grounding as well as complex reasoning tasks such as QA in driving scenarios. 

We evaluate our model's performance in both 3D grounding and 3D complex reasoning tasks on various indoor and outdoor datasets as well as a standard \mmfm{} benchmark, and show qualitative results in 3D grounding in non-driving scenes (Fig.~\ref{fig:vis}).  
For 3D grounding on the \ttc dataset~\cite{deruyttere2019talk2car}, \ours surpasses the baselines by \textbf{21.3} in Bird's Eye View (BEV) AP (71.4 vs 50.1) and by \textbf{18.7} in 3D AP (64.1 vs 45.4).
Additionally, our training framework improves the performance of \ours on the \dLM~\cite{sima2023drivelm} dataset, nearly doubling the performance in the BEV AP (66.0 vs 33.2) for 3D grounding from a baseline. 
We also test \ours on complex reasoning benchmark of driving scenarios (\dLM), and improve the overall score by \textbf{17.7} (50.1 vs 32.4) compared to \dLM baseline~\cite{sima2023drivelm}.
Furthermore, we show that \ours performs the state-of-the-art in 2D referring expression comprehension, achieving the average score of \textbf{87.0} on refCOCO/$\texttt{+}$/g.
Finally, we show that \ours maintains competitive performance in various MLLM benchmarks including VQAv2, GQA, etc., confirming that our 3D reasoning capability is an \emph{expansion}, not a \textit{trade-off}.

\section{Related Work}
\label{sec:related_work}

\myparagraph{Vision Language Models.}
By scaling up pre-training on the internet-scale dataset, there has been significant progress for VLMs in the 2D vision-language domain, showing strong capabilities in few-shot generalization. VLBRRT~\cite{vlbert} and ViLBERT~\cite{lu2019vilbert} capitalized on a BERT-style framework for image-text co-embedding. CLIP~\cite{radford2021learning} embedded images and text captions into a shared feature space via contrastive learning and pioneered zero-shot vision tasks. BLIP~\cite{li2022blip} and BLIP2~\cite{Li2023BLIP2BL} further improved CLIP by leveraging extra pseudo-labeled data and better image/language encoders. Flamingo~\cite{Alayrac2022FlamingoAV} and its open-source implementation Open-Flamingo~\cite{Awadalla2023OpenFlamingoAO} proposed a fast adaptation approach to enable in-context few-shot learning on novel visual-language tasks. GPT4V~\cite{Achiam2023GPT4TR} and Gemini~\cite{team2023gemini} further demonstrated state-of-the-art human-level visual reasoning ability through scaling. LLaVA~\cite{liu2023llava} pioneered instruction fine-tuning in the multimodal field. 
These works have predominantly focused on the 2D vision and language tasks. On the other hand, we aim to adapt these \mmfm{}s to enhance their capabilities for complex 3D reasoning and scene understanding tasks.

\myparagraph{{Image-grounded Reasoning.}}
With the advancement of multi-modal large language models, image-grounded reasoning (referring and grounding) has shown a great progress in 2D space.
Image-grounded reasoning requires a model to localize an object or a region that an input prompt enquires, or describe about a region of interest. 
VisionLLM~\cite{wang2024visionllm} adapts 2D object detector to align with an LLM, and GPT4-ROI~\cite{zhang2023gpt4roi} employs hierarchical feature modeling of detectors to reason about input visual prompt (ROI). 
Kosmos-2~\cite{peng2023kosmos} and Shikra~\cite{chen2023shikra} have shown pure transformer-based visual encoder can surpass using 2D detectors with data scaling. 
Recently, Ferret~\cite{you2023ferret} has shown remarkable image-grounded reasoning from both free-form visual prompt and text prompt. 
These works reason in 2D space (image coordinate).
To the best of our knowledge, our work is the first to expand the reasoning capability of a MLLM to 3D.

\myparagraph{Reasoning with Language Models in Autonomous Driving.} 
Reasoning is a long-standing problem in autonomous driving. GPT-Driver~\cite{mao2023gpt} incorporated reasoning into autonomous driving by reformulating motion planning as a language modeling problem and introducing fine-tuned LLMs as a motion planner. DriveGPT4~\cite{xu2023drivegpt4} proposed an end-to-end driving approach that leverages a vision language model to directly predict actions from sensory inputs. DiLu~\cite{wen2023dilu} capitalized on a knowledge-driven approach with large language models to improve reasoning capability. Agent-Driver~\cite{mao2023language} proposed an LLM-based embodied agent, achieving state-of-the-art decision-making performance and making it transparent. Dolphins~\cite{ma2023dolphins} proposed an instruction fine-tuning pipeline for \mmfm{}-based autonomous driving. DriveLM~\cite{sima2023drivelm} opted to use graph visual question answering to tackle reasoning problems in autonomous driving. LMDrive~\cite{shao2023lmdrive} tackled closed-loop autonomous driving with language models. Our method is closely related to these prior works since we attempt to solve visual reasoning problems in autonomous driving. In contrast to prior work, \ours{} can directly reason in the 3D space for complex AV perception scenarios, and can be trained in an end-to-end fashion.

\section{Unified Language-Image Pretraining for 2D and 3D}
\label{sec:method}

Our goal is to expand the capabilities of vision-language models to reason in 3-dimensional space. 
We propose a unified training framework to learn from both 2D and 3D perceptual data as well as standard image-text pairs. 
In this section, we first discuss the data standardization to train a vision-language model at scale (Sec.~\ref{sec:data_scale}), {task scaling} to understand perceptual information in versatile I/O format (Sec.~\ref{sec:task_scale}), \textit{visual chain-of-thought} reasoning for 3D grounding and question answering tasks (Sec.~\ref{sec:v_cot}), and finally, we present \ours, the final model of our unified training framework built on LLaVA-1.5~\cite{liu2023improvedllava} (Sec.~\ref{sec:arch}).

\subsection{Data-scaling for Image-based 3D Reasoning}
\label{sec:data_scale}

Our goal is to train a single 2D + 3D MLLM from all data sources  available. To standardize many different 2D and 3D grounding tasks into one, we standardize the data, phrase all tasks as next token prediction, and format 3D reasoning as a multi-turn conversation.

\vspace{0.2cm}\myparagraph{Data standardization.} 
We consider points and boxes as our main spatial representation for 2D and 3D reasoning. 
We convert every label to either a point $o^{\text{2D}}_{\text{point}}=[\hat{x}, \hat{y}]$ or a bounding box $o^{\text{2D}}_{\text{box}}=[\hat{x}, \hat{y}, \hat{x}', \hat{y}']$.
Similarly, we convert every 3D label to either 
$o^{\text{3D}}_{\text{point}}=[{x}, {y}, z]$ or  $o^{\text{3D}}_{\text{box}}=[{x}, {y}, z, w, h, l, r_1, r_2, r_3]$ where $r_1$, $r_2$, $r_3$ are Euler angles. 
We first standardize image-based 3D datasets by unifying camera parameters. 
We follow the procedure of Omni3D~\cite{brazil2023omni3d}; define a virtual camera with a fixed focal length $f$ and transform depth $z$ according to the original camera parameters and the target image size. 
Since all 3D labels are unified to a consistent camera intrinsic, we can now convert all x and y coordinates to 2D projected coordinates ($\hat{x}, \hat{y}$). 
Consequently, we can represent all label formats to naturally follow 2D to 3D per-object token sequence: 
\begin{align}
    o^{\textcolor{mytangoblue}{\textbf{2D}}}_{\text{point}} &= [\textcolor{mytangoblue}{\hat{x}}, \textcolor{mytangoblue}{\hat{y}}] \\
    o^{\textcolor{mytangoblue}{\textbf{2D}}}_{\text{box}}\ \;   &= [\textcolor{mytangoblue}{\hat{x}}, \textcolor{mytangoblue}{\hat{y}}, \textcolor{mytangoblue}{\hat{x}'}, \textcolor{mytangoblue}{\hat{y}'}] \\
    o^{\textcolor{myred}{\textbf{3D}}}_{\text{point}} &= [\textcolor{mytangoblue}{\hat{x}}, \textcolor{mytangoblue}{\hat{y}}, \textcolor{myred}{{z}}] \\
    o^{\textcolor{myred}{\textbf{3D}}}_{\text{box}} \ \;  &= [\textcolor{mytangoblue}{\hat{x}}, \textcolor{mytangoblue}{\hat{y}}, \textcolor{myred}{{z}}, \textcolor{myred}{{w}}, \textcolor{myred}{{h}}, \textcolor{myred}{{l}}, \textcolor{myred}{{r}_1}, \textcolor{myred}{{r}_2}, \textcolor{myred}{{r}_3}]
\end{align}
where each value is represented as a short sequence of text tokens (3 for 3-decimal numbers). 
This allows the model to predict consistent ordering of token sequence from 2D to 3D, which improves the understanding of the underlying structure.
With autoregressive models, we first localize objects in image coordinates ($\hat{x}, \hat{y}$), then infer depth ($z$), and then infer the size and orientation ($w,h,l,r_1,r_2,r_3$).

\vspace{0.2cm}\myparagraph{3D reasoning as multi-turn conversations.}
Now, we combine the 2D and 3D data with language-image instruction tuning data of visual language models~\cite{liu2023llava}.
For each image and a set of object labels pair, we construct a multi-turn conversational question-answer data (\textbf{Q}$_1$, \textbf{A}$_1$, \textbf{Q}$_2$, \textbf{A}$_2$, $\dots$, \textbf{Q}$_n$, \textbf{A}$_n$). 
Each question refers an object with one property $b_q$ and enquires $b_a$:
\begin{align}
b_q, b_a \in \{\text{box}_\text{2D}, \text{caption}, \text{box}_\text{3D}\} 
\end{align} 
Each object property has a set of {prompts} predefined, such as \textcolor{darkgrey}{\texttt{``Provide the 3D bounding box of the region this sentence describes: <caption>''}} for $b_q=\text{caption}$ and $b_a=\text{box}_\text{3D}$.
We combine the meta information of objects (e.g., attribute, physical state, etc.) with the class name to enrich the textual information.

\subsection{Task-scaling for Versatile I/O Format}
\label{sec:task_scale}

\begin{figure}[!t]
    \centering
    \includegraphics[width=\linewidth]{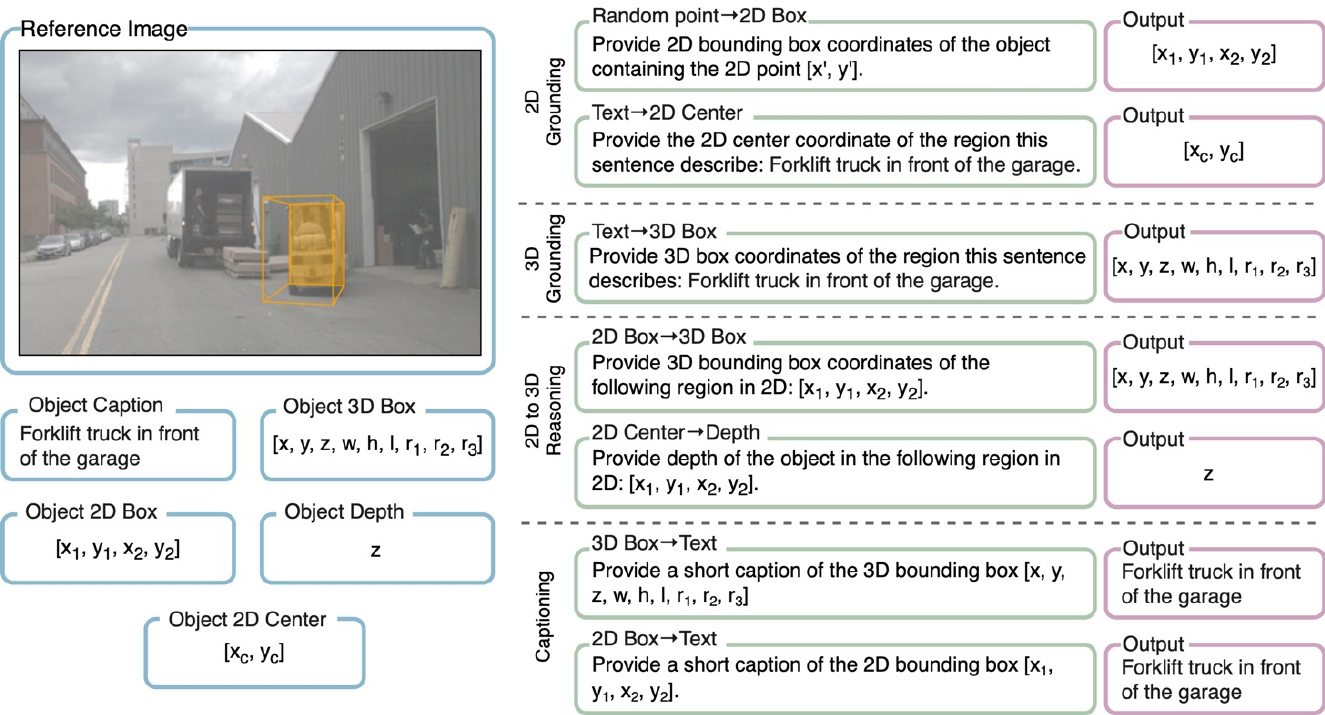}
    \caption{\textbf{Task-scaling for versatile I/O format.} Decomposing the existing label formats for 3D grounding task. A complete 3D location can be decomposed into a center point (\textcolor{darkgrey}{\texttt{[x, y, z]}}), a depth (\textcolor{darkgrey}{\texttt{[z]}}), a (projected) 2D point (\textcolor{darkgrey}{\texttt{[x$_\text{c}$, y$_\text{c}$]}}), and a (projected) 2D box (\textcolor{darkgrey}{\texttt{[x1, y1, x2, y2]}}). We define various tasks that connect among these to train versatile I/O formats. \textbf{Left}: available (decomposed) annotations. \textbf{Right}: various tasks for training. }
    \label{fig:tasks}
    \vspace{-0.5cm}
\end{figure}

\begin{figure}[!t]
    \centering
    \includegraphics[width=\linewidth]{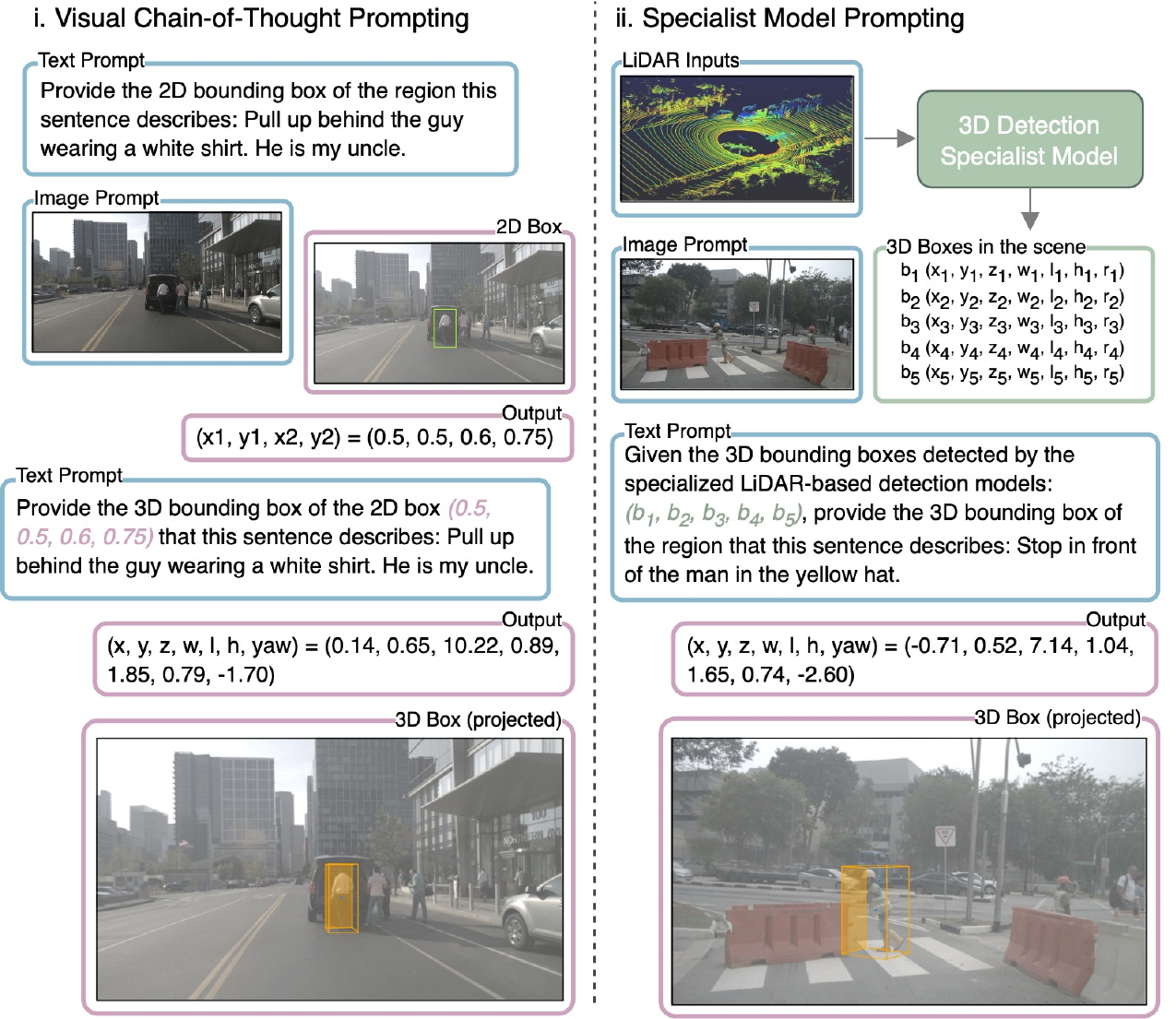}
    \vspace{-0.5cm}
    \caption{\textbf{\ours inference with prompting}. \textbf{Left}: Visual Chain-of-Thought Prompting to reason in 3D step-by-step. \textbf{Right}: Incorporating specialist models to further improve localization of \ours. 
    Our model can either predict directly from text prompt, or with visual chain-of-thought prompting, or with specialist predictions as prompt. 
    }
    \label{fig:method}
    \vspace{-0.8cm}
\end{figure}

We are interested in a generalist model that accepts input and generates     output in versatile formats.
Users may want to supplement 2D points or boxes as visual prompt during inference, or may only want the metric depth of an object instead of a complete 3D location. 
This interest in versatile I/O format shares the same spirit of instruction tuning in 2D-based visual language models~\cite{liu2023llava,Dai2023InstructBLIPTG,Alayrac2022FlamingoAV}. 
To this end, we define multiple relevant tasks for a model to adapt to wider spectrum of similar tasks in 2D and 3D. 
We start by \emph{decomposing} the existing label formats to easier tasks as illustrated in Figure~\ref{fig:tasks}. 
After, we have expanded set of object properties to construct question-answer pairs:
\begin{align}
b_q &\in \{ \text{point}_\text{2D}, \text{box}_\text{2D}, \text{caption}, \text{point}_\text{3D}, \text{box}_\text{3D} \} \\
b_a &\in \{\text{point}_\text{2D}, \text{box}_\text{2D}, \text{caption}, \text{depth}, \text{point}_\text{3D}, \text{box}_\text{3D} \} 
\end{align}
We construct up to $n=30$ question answer pairs $(\textbf{Q}^{b_q}_{b_a},\textbf{A}_{b_a})$ sampled at random for each data. 
We combine a collection of 2D and 3D vision datasets (\dataset{}), summarized in Table~\ref{tab:dataset_summary}, and jointly train with this expanded set of tasks.

\subsection{Visual Chain-of-Thought Prompting}
\label{sec:v_cot}

One of the most intriguing properties of large language models is its \emph{emergent} ability to improve reasoning with intermediate steps~\cite{wei2022chain}.
This mostly attributes to vast corpus of rich text data with numerous step-by-step question answering samples~\cite{wei2022finetuned}. 
We artificially supplement this \emph{step-by-step} reasoning of 3D by interleaving multiple questions of the same object from easy-to-hard order (the left part of Figure.~\ref{fig:method}): 
\begin{align}
\text{maximize} \quad 
\begin{cases}\quad 
    p(\text{\textbf{A}}_{\text{box}_\text{2D}}|\textbf{Q}^{\text{caption}}_{\text{box}_\text{2D}}) & \textit{question 1} \\ \quad p(\text{\textbf{A}}_{\text{box}_\text{3D}}|\textbf{Q}^{\text{caption}}_{\text{box}_\text{2D}},\text{\textbf{A}}_{\text{box}_\text{2D}},\textbf{Q}^{\text{caption}}_{\text{box}_\text{3D}}) & \textit{question 2} \\
    \quad \quad ...
\end{cases}
\end{align}
Furthermore, we allow test-time adaptation to any specialist models by mixing in \emph{candidate} objects as a system prompt (the right part of Figure.~\ref{fig:method}). 
This effectively alleviates the problem of localizing in 3D to ``choosing the appropriate box from candidates'',
\begin{align}
\text{maximize} \quad p(\text{\textbf{A}}_{\text{box}_\text{3D}}|\textbf{S}_{\text{box}_{\text{3D}}}, \textbf{Q}^{\text{caption}}_{\text{box}_\text{3D}})
\end{align}
where \textbf{S}$_{\text{box}_{\text{3D}}}$ is a set of candidate boxes, which can be provided by any specialist models (depending on available input modalities) at inference. 
During training, we use the ground truth boxes with a prompt \textcolor{specialist_green}{\texttt{``Here is the list of 3D bounding boxes of all objects around the camera:''}} and our model does not bind with any particular specialist model. 

\subsection{\ours}
\label{sec:arch}

We introduce \ours, a multi-modal large language model based on LLaVA-1.5 architecture trained to reason in both 2D and 3D. 
Although we maintain the generality of model architecture, we make simple yet critical changes to the original LLaVA. 
We first replace the CLIP visual encoder with DINOv2~\cite{oquab2024dinov}, and undergo the same alignment step of the original LLaVA. 
Although DINOv2 is not a text-aligned visual encoder like CLIP, we found minimal degradation in the standard visual language model benchmarks while significantly improving 3D-related tasks. 
Then, we finetune the language model (Vicuna-7B~\cite{vicuna2023}) while freezing the visual encoder and jointly on LLaVA instruction-following data and the 2D part of LV3D following Sec.~\ref{sec:data_scale},~\ref{sec:task_scale} and~\ref{sec:v_cot}. 
We use low image resolution ($336 \times 336$) and train with a large batch size. 
Then, we add an additional finetuning stage for both visual and language models with high resolution images ($672 \times 672$) of the full LV3D. 
For 3D representation, we use log-scale for depth and all others remain unchanged. 
For 2D, we normalize image coordinates between 0 and 999.
For 3D, we we filter-out all annotations outside $X_{\text{min}}$ and $X_{\text{max}}$ (different for each $\hat{x}, \hat{y}, z, w, h, l, r_1, r_2, r_3$, more in supplementary), and normalize between 0 and 999.

\section{Experiments}
\label{sec:expts}
We evaluate the effectiveness of \ours in three aspects: (1) 3D-grounded reasoning for indoor and outdoor scenes, (2) complex reasoning in 3D, and (3) standard vision-language benchmarks. 
\subsection{Implementation Details}
We use LLaVA-1.5 \cite{liu2023improvedllava} with Vicuna-7B as our base model. 
We replace the CLIP visual encoder with ViT-L/14~\cite{dosovitskiy2020vit} based DINOv2. 
For all localization outputs, we use 3 decimal places with text tokens, and keep 3 tokens per value (e.g., \textcolor{darkgrey}{\texttt{[021,521, ...]}}). 
Accordingly, we pre-process all LLaVA instruction-following data to reflect this change. 
We follow the same alignment step to train the MLP projection layers with the same training setup in~\cite{liu2023improvedllava}. 
For 2D and 3D pretraining, we use random sampling following the sampling rate in Table~\ref{tab:dataset_summary}. 
Each data sample (image-annotation pair) is converted to the multi-turn conversation format (Fig.~\ref{fig:tasks}) sampled at random. 
During pretraining, we use 8$\times$8 A100s with a batch size of 1024 and train the model with a learning rate $lr=2\times 10^{-5}$ on images with $336 \times 336$ resolution. 
Then, we fine-tune all parameters including the visual encoder on a higher resolution $672 \times 672$ with 8$\times$8 A100s and a batch size of 256 with 4 gradient accumulation steps (effective batch size of 1024) and a learning rate $lr=2\times 10^{-5}$.

\begin{table}[t]
\centering
\caption{\textbf{2D and 3D Language-Image Pretraining Dataset (\dataset)}. Summary of components detailing the number of images, tasks, availability of 2D and 3D labels, the number of QAs and objects, and their multiples during training (\textit{stage 1} and \textit{stage 2}). $\star$: Only used 2D bounding box. 
}
\resizebox{0.85\textwidth}{!}{
\begin{tabular}{llcccrrr}
\toprule
{dataset} & {images} & \ {labels}$_{2D}$ & \ {labels}$_{3D}$ & captions & \ {\# QAs }\ & stage 1 & stage 2 \\
\midrule
LLaVA data~\cite{liu2023llava} & 80K &  \checkmark & \xmark & \checkmark & 158K  & 1 & 0.5 \\
refCOCO/\texttt{+}/g~\cite{refcoco} & 67K &  \checkmark & \xmark & \checkmark & 154K & 1 & 0.5 \\
GRIT (subset)~\cite{peng2023kosmos} & 4M &  \checkmark & \xmark & \checkmark & 6.9M & 1 & 0.3 \\
AS (filtered)~\cite{wang2024allseeing_v2} & 3.7M & \checkmark & \xmark & \checkmark & 13.2M & 1 & 0.5  \\
COCO~\cite{lin2014microsoft} & 118K & \checkmark & \xmark & \xmark & 860K & 1 & 0.5 \\
Objects365~\cite{o365} & 600K &  \checkmark & \xmark & \xmark  & 25.4M & 0.3 & 0.2 \\
SUN-RGBD~\cite{song2015sun} & 5K & \checkmark & \checkmark & \xmark & 41K & 1$^\star$ & 5 \\
Hypersim~\cite{hypersim} & 67K & \checkmark & \checkmark & \xmark & 2M & 1$^\star$ & 5 \\
ArkitScenes~\cite{dehghan2021arkitscenes} & 53K & \checkmark & \checkmark & \xmark & 420K& 1$^\star$ & 5 \\
Objectron~\cite{objectron2021} & 37K & \checkmark & \checkmark & \xmark & 43K & 1$^\star$ & 5 \\
KITTI~\cite{Geiger2012CVPR} & 4K & \checkmark & \checkmark & \xmark & 25K & 1$^\star$ & 5 \\
NuScenes~\cite{nuscenes2019} & 40K &  \checkmark & \checkmark & \xmark &  1.1M & 1$^\star$ & 2 \\
Lyft~\cite{houston2021one} & 105K & \checkmark & \checkmark & \xmark & 723K & 0 & 2 \\
Argoverse2~\cite{Argoverse2} & 79K & \checkmark & \checkmark & \xmark & 915K & 0 & 4 \\
Waymo~\cite{waymo} & 680K & \checkmark & \checkmark & \xmark & 5.1M & 0 & 0.4 \\
\midrule
Total & 9.6M &  \checkmark & \checkmark & \checkmark & 40.9M & 0.87 & 0.52 \\
\bottomrule
\end{tabular}
}
\vspace{-0.5cm}
\label{tab:dataset_summary}
\end{table}

\subsection{Datasets}

We pre-train \ours on \dataset{}, and then fine-tune it on the training split of the target datasets, \ttc and \dLM. 

\myparagraph{\ttc}~\cite{deruyttere2019talk2car} is a 3D referring expression comprehension dataset of various driving scenarios. 
It consists of 8,349 training samples and 1,163 validation samples with images and LiDAR data. 
It provides rich question-answer pairs grounded to an object in the image. 
Each object is labeled with a situational text that uniquely identifies the object (e.g., ``\textit{Wow hold on! That looks like my stolen bike over there! Drop me off next to it.}''). 
The original benchmark~\cite{deruyttere2019talk2car} evaluates the 2D grounding performance with the AP$_{0.5}$ metric. 
MSSG~\cite{cheng2023language} extends the task to 3D grounding and evaluates on both BEV AP and 3D AP.

\myparagraph{\dLM}~\cite{sima2023drivelm} is a recently released question-answering dataset for autonomous driving based on the nuScenes dataset~\cite{nuscenes2019}.
It consists of various driving scenes with multi-view images and LiDAR point clouds, as well as frame-level question-answering data, and has a total of 4,871 frames. 
Each frame contains 91.4 question-answer pairs on average, covering core autonomous driving tasks such as perception, prediction, and planning, as well as a short description and 2D boxes of important objects.  
To evaluate \ours, we construct another 3D grounding benchmark based on the \dLM dataset, which we call \textbf{\dLM-Grounding}.
We associate the 2D boxes with the nuScenes 3D bounding boxes by computing the IoU between 2D boxes projected from 3D labels and the 2D boxes of labeled important objects, and only keep those with a IoU greater than $0.35$. 
After association, \dLM-Grounding has a total of 13,287 images, about one annotation per image. 
We also use the \textbf{\dLM-QA} data from the original {\dLM} to fine-tune \ours for complex reasoning tasks.
The original training split has 696 scenes in total. 
We use 600 scenes for training and 96 scenes for validation, which we include the \dLM provided scenes for sample evaluation and \ttc validation split scenes. 
We evaluate 3D grounding with the same BEV AP and 3D AP metric as those in \ttc.

\subsection{3D-Grounded Reasoning}

\begin{table*}[!t]
\centering
\caption{\textbf{\ttc Benchmark for 2D and 3D Grounding}. We denote C as Camera and L as LiDAR. $\dagger$: we use the top-30 predicted boxes of CenterPoint~\cite{yin2021center} as visual prompt as illustrated in Figure~\ref{fig:method}. AP$_\text{A}$ and AP$_\text{B}$ follow MSSG~\cite{cheng2023language} that apply different IoU threshold for each category.
}
\resizebox{0.6\textwidth}{!}{
\begin{tabular}{lcccccc}
\toprule
\multicolumn{1}{c}{\multirow{2}{*}{Method}} & \multicolumn{1}{c}{\multirow{2}{*}{Input}} & 2D & \multicolumn{2}{c}{BEV} & \multicolumn{2}{c}{3D} \\
  & & AP$_{0.5}$ & AP$_{\text{A}}$ & AP$_{\text{B}}$ & AP$_{\text{A}}$ & AP$_{\text{B}}$ \\
\midrule
\textit{2D Specialist} & & & & & \\
\ttc-2D~\cite{deruyttere2019talk2car} & C & 50.5 & - & - & - & - \\
VL-Bert~\cite{vlbert} & C & 63.1 & - & - & - & - \\
ViLBERT~\cite{lu2019vilbert} & C & 68.9 & - & - & - & - \\
CMRT~\cite{cmrt} & C & 69.1 & - & - & - & - \\
Stacked VLBert~\cite{stacked_vlbert} & C & 71.0 & - & - & - & - \\
FA~\cite{fa} & C & 73.5 & - & - & - & - \\
\midrule 
\ours (zero-shot) & C & 46.3 & 32.0 & 19.5 & 22.3 & 9.8 \\
\ours & C & \textbf{79.2} & 46.3 & 30.1 & 34.7 & 18.2 \\
\midrule
\textit{3D Specialist} & & & & \\
\ttc-3D~\cite{deruyttere2019talk2car} & L + C & - & 30.6 & 24.4 & 27.9 & 19.1 \\
MSSG~\cite{cheng2023language} & L + C & - & 50.1 & 35.7 & 45.4 & 23.7 \\
\midrule
\ours$^\dagger$ & L + C & 76.3 & \textbf{71.4}& \textbf{61.2} & \textbf{64.1} & \textbf{39.8}  \\
\bottomrule
\end{tabular}
}
\label{tab:talk2car_main}
\vspace{-0.5cm}
\end{table*}
\begin{table*}[!t]
\centering
\caption{
\textbf{\dLM-Grounding benchmark for 3D grounding.} LV3D (2D) indicates that only 2D data in the pre-train dataset is included. 
We finetune \ours and LLaVA-1.5~\cite{liu2023improvedllava} on the \dLM-Grounding dataset. 
\ours pre-trained with \dataset achieves a 99\% improvement compared to LLaVA-1.5 on the AP$_{\text{A}}^{\text{BEV}}$ metric. 
}
\resizebox{0.6\textwidth}{!}{
\begin{tabular}{llcccc}
\toprule
\multicolumn{1}{c}{\multirow{2}{*}{Method}} & \multicolumn{1}{c}{\multirow{2}{*}{Pre-train Data}} & \multicolumn{2}{c}{BEV} & \multicolumn{2}{c}{3D} \\
  &  & AP$_{\text{A}}^{\text{BEV}}$ & AP$_{\text{B}}^{\text{BEV}}$ & AP$_{\text{A}}^{\text{3D}}$ & AP$_{\text{B}}^{\text{3D}}$ \\
\midrule
LLaVA-1.5~\cite{liu2023improvedllava} & LLaVA data & 33.2 & 16.3 & 21.7 & 7.7 \\
\ours & LLaVA data & 39.6 & 21.7 & 25.8 & 10.5 \\ 
\ours & LV3D (2D) & 50.5 & 31.2 & 32.5 & 17.3 \\ 
\ours & LV3D & \textbf{66.0} & \textbf{52.1} & \textbf{56.2} & \textbf{40.5} \\ 
\bottomrule
\end{tabular}
}
\label{tab:drivelm_main}
\end{table*}
\begin{table*}[!t]
\centering
\caption{\textbf{Inodoor 3D Grounding Benchmark.} Here we compare \ours trained on ``small'' subset of LV3D and the full LV3D. Although the subset and full LV3D share the same indoor datasets, the added 2D data and outdoor 3D data translate to better indoor 3D grounding result. 
}
\resizebox{0.9\textwidth}{!}{
\begin{tabular}{l>{\centering\arraybackslash}m{1.6cm}>{\centering\arraybackslash}m{1.6cm}>{\centering\arraybackslash}m{1.6cm}>{\centering\arraybackslash}m{1.6cm}>{\centering\arraybackslash}m{1.6cm}>{\centering\arraybackslash}m{1.6cm}}
\toprule
\multicolumn{1}{c}{\multirow{2}{*}{Pre-train Data}} & \multicolumn{2}{c}{Objectron~\cite{objectron2021}} & \multicolumn{2}{c}{ArkitScenes~\cite{dehghan2021arkitscenes}}  & \multicolumn{2}{c}{SUN-RGBD~\cite{song2015sun}} \\
 & mAP$^\text{cls}_\text{3D}$ &mAP$^\text{cls+loc}_\text{3D}$ & mAP$^\text{cls}_\text{3D}$ &mAP$^\text{cls+loc}_\text{3D}$ & mAP$^\text{cls}_\text{3D}$ &mAP$^\text{cls+loc}_\text{3D}$    \\
\midrule 
LV3D-small & 56.7 & 36.1 & 21.6 & 28.3 & 25.5 & 25.5 \\
LV3D       & 69.8 & 45.4 & 23.5 & 31.8 & 29.7 & 28.8 \\
$\Delta$ & \textbf{13.1} & \textbf{9.3} & \textbf{1.9} & \textbf{3.5} & \textbf{4.2} & \textbf{3.3} \\
\bottomrule
\end{tabular}
}
\label{tab:omni3d}
\end{table*}
\begin{table*}[!t]
\centering
\vspace{-0.5cm}
\caption{\textbf{Referring Expression Comprehension Benchmark}. We compare \ours with other MLLMs for general 2D grounding tasks. \ours consistently performs best in all data splits in refCOCO.  
}
\resizebox{0.8\textwidth}{!}{
\begin{tabular}
{lc|ccc|ccc|cc|c}
\toprule
\multicolumn{1}{c}{\multirow{2}{*}{Models}} & \multicolumn{1}{c}{\multirow{2}{*}{Size}} & \multicolumn{3}{c}{RefCOCO} & \multicolumn{3}{c}{RefCOCO\texttt{+}}  & \multicolumn{2}{c}{RefCOCOg} & \\
  & & val & testA & testB & val & testA & testB & val & test & Avg. \\
\midrule
\textit{Specialist} & & & & &&&&&& \\
MAttNet~\cite{yu2018mattnet} & & 76.4 & 80.4 & 69.3 & 64.9 & 70.3 & 56.0 & 66.7 & 67.0 & 68.9 \\ 
OFA-L~\cite{wang2022ofa} & & 80.0 & 83.7 & 76.4 & 68.3 & 76.0 & 61.8 & 67.6 & 67.6 & 72.7 \\
TransVG~\cite{Deng2021TransVGEV} & & 81.0 & 82.7 & 78.4 & 64.8 & 70.7 & 56.9 & 68.7 & 67.7 & 71.4 \\
UNITER~\cite{chen2020uniter} & & 81.4 & 87.0 & 74.2 & 75.9 & 81.5 & 66.7 & 74.0 & 68.7 & 76.2 \\
VILLA~\cite{villa} & & 82.4 & 87.5 & 74.8 & 76.2 & 81.5 & 66.8 & 76.2 & 76.7 & 77.8 \\
UniTAB~\cite{yang2022unitab} & & 86.3 & 88.8 & 80.6 & 78.7 & 83.2 & 69.5 & 80.0 & 80.0 & 80.6\\
MDETR~\cite{kamath2021mdetr} & & 86.8 & 89.6 & 81.4 & 79.5 & 84.1 & 70.6 & 81.6 & 80.9 & 81.8 \\
\midrule 
\textit{Generalist} &&&&&&&&&&\\
LLaVA-1.5~\cite{liu2023improvedllava} & 7B & 75.6 & 82.1 & 66.9 & 65.5 & 76.2 & 53.9 & 68.9 & 69.1 & 69.8  \\
VisionLLM-H~\cite{wang2024visionllm} & 7B & 86.7 & - & - & - & - & - & - & - & -  \\
Shikra~\cite{chen2023shikra} & 7B & 87.0 & 90.6 & 80.2 & {81.6} & {87.4} & 72.1 & 82.3 & 82.2 & 82.9 \\
Ferret~\cite{you2023ferret} & 7B & 87.5 & 91.4 & 82.5 & 80.8 & {87.4 }& 73.1 & 83.9 & {84.8} & 83.9 \\
MiniGPT-v2~\cite{chen2023minigptv2} & 7B & 88.7 & 91.7 & 85.3 &  80.0 & 85.1 & 74.5 & 84.4 & 84.7 & 83.8 \\
LLaVA-G~\cite{zhang2023llavagrounding} & 7B & 89.2 & - & - & 81.7 & - & - & 84.8 & - & - \\
Qwen-VL~\cite{Qwen-VL} & 7B & 88.6 & 92.3 & 84.5 & 82.8 & 88.6 & 76.8 & 86.0 & 86.3 & 85.7  \\
\ours & 7B & \textbf{90.9} & \textbf{92.6} & \textbf{87.9} & \textbf{83.9} & \textbf{89.2} & \textbf{77.4} & \textbf{86.6} & \textbf{87.2} & \textbf{87.0} \\
\bottomrule
\end{tabular}
}
\label{tab:rec}
\end{table*}
\begin{table*}[!t]
\centering
\vspace{-0.5cm}
\caption{\textbf{MLLM Benchmarks}. We compare \ours with other MLLMs for various visual question-answering tasks. 
}
\resizebox{0.82\textwidth}{!}{
\begin{tabular}{lccccccc}
\toprule
Model & Size & VQA$^\text{v2}$~\cite{goyal2017making} & GQA~\cite{hudson2019gqa} & VizWiz~\cite{gurari2018vizwiz} & SQA$^\text{I}$~\cite{lu2022learn} & POPE~\cite{Li-hallucination-2023} \\ 
\midrule 
BLIP-2~\cite{Li2023BLIP2BL} & 13B & 41.0 & 41.0 & 19.6 & 61.0 & 85.3 \\
InstructBLIP~\cite{Dai2023InstructBLIPTG} & 7B & - & 49.2 & 34.5 & 60.5 & - \\
InstructBLIP~\cite{Dai2023InstructBLIPTG} & 13B & - & 49.5 & 33.4 & 63.1 & 78.9 \\
IDEFICS~\cite{laurencon2023obelics} & 9B & 50.9 & 38.4 & 35.5 & - & - \\ 
\midrule 
Shikra~\cite{chen2023shikra} & 13B & 77.4 & - & - & - & -  \\
Qwen-VL~\cite{Qwen-VL} & 7B & \textbf{78.8} & 59.3 & 35.2 & 67.1 & - \\
Qwen-VL (chat)~\cite{Qwen-VL} & 7B & 78.2 & 57.5 & 38.9 & 68.2 & -  \\
miniGPT-v2~\cite{chen2023minigptv2} & 7B & - & 60.1 & \textbf{53.6} & - & -   \\
LLaVA-1.5~\cite{liu2023improvedllava} & 7B & 78.5 & 62.0 & 50.0 & 66.8 & 85.9  \\
\ours & 7B & 78.3 & \textbf{62.4} & 51.0 & \textbf{69.2} & \textbf{87.1}  \\
\bottomrule
\end{tabular}
}
\label{tab:mllm}
\end{table*}

\begin{table}[ht]
\centering
\caption{\textbf{Improvements of data scaling}. This table illustrates how the inclusion of different datasets affects performance on \ttc 3D grounding benchmark. Note that we do not perform visual chain-of-thought prompting nor high-resolution finetuning for better comparison. We evaluate the model on image resolution $336 \times 336$. 
}
\resizebox{0.9\textwidth}{!}{
\begin{tabular}{cccccccccccc}
\toprule
LLaVA & refC. & COCO & \ttc & Nus. & O365 & \dLM & Omni3D & AP$_{\text{A}}^{\text{BEV}}$ & AP$_{\text{B}}^{\text{BEV}}$ & AP$_{\text{A}}^{\text{3D}}$ & AP$_{\text{B}}^{\text{3D}}$ \\
\midrule
\checkmark &  &  &  &  &  & &  & 19.7 & 9.4 & 10.3 & 3.3 \\
\checkmark & \checkmark &  &  &  &  &  &  & 25.9 & 11.3 & 12.1 & 3.9 \\
\checkmark & \checkmark & \checkmark &  &  &  &  &  & 27.0 & 14.0 & 14.8 & 5.7 \\
\checkmark & \checkmark & \checkmark & \checkmark & \checkmark &  &  &  & 33.6 & 20.8 & 23.4 & 13.3 \\
\checkmark & \checkmark & \checkmark & \checkmark & \checkmark & \checkmark & \checkmark & & 42.4 & 27.7 & 31.0 & 16.2 \\
\checkmark & \checkmark & \checkmark & \checkmark & \checkmark & \checkmark & \checkmark & \checkmark & \textbf{44.7} & \textbf{28.7} & \textbf{32.0} & \textbf{16.5} \\
\bottomrule
\end{tabular}
}

\label{tab:dataset_performance}
\end{table}

\begin{table}[ht]
\centering
\caption{\textbf{Visual Chain-of-Thought ablations.} We evaluate \ours with and without visual Chain-of-thoughts (VCoT) prompting during inference on the \ttc dataset for the 3D grounding task. \label{tab:vcot_ablation} 
}
\resizebox{0.35\textwidth}{!}{
\begin{tabular}{cccccc}
\toprule
VCoT & AP$_{\text{A}}^{\text{BEV}}$ & AP$_{\text{B}}^{\text{BEV}}$ & AP$_{\text{A}}^{\text{3D}}$ & AP$_{\text{B}}^{\text{3D}}$ \\ 
\midrule
                     & 43.6 & 28.2 & 32.7 & 15.9 \\
  \checkmark        & \textbf{46.3} & \textbf{30.1} & \textbf{34.7} & \textbf{18.2} \\
\bottomrule
\end{tabular}
}
\end{table}

Our results for 3D grounding on the \ttc dataset are detailed in Table \ref{tab:talk2car_main}, which is structured according to the input modalities used for 3D grounding. 
The baselines that rely solely on camera inputs are only evaluated on 2D grounding, whereas those incorporating both camera and LiDAR inputs are evaluated on both 2D and 3D grounding. \ours is pre-trained on \dataset{} and fine-tuned on \ttc with resolution $672 \times 672$. We apply visual Chain-of-Thought when predicting the 3D grounding.
Remarkably, our camera-only \ours significantly surpasses the state-of-the-art model FA~\cite{fa} by 5.7 points on 2D AP$_{0.5}$. Surprisingly, \ours also outperforms the camera+LiDAR baseline, \ttc-3D~\cite{deruyttere2019talk2car}, by 15.7 points on the BEV AP$_\text{A}$ metric. 
Our camera-only \ours is only 3.8 points behind the state-of-the-art camera+LiDAR baseline MSSG~\cite{cheng2023language}. 
MSSG~\cite{cheng2023language} utilized the LiDAR point encoder similar to CenterPoint~\cite{yin2021center} as well as image and text encoders for predicting 3D grounding. Similarly, we leverage the LiDAR modality by using the predictions from CenterPoint~\cite{yin2021center}, selecting the top-30 boxes with the highest confidence as a visual prompt.
We observe a substantial 25.1 points improvements in AP$_\text{A}$, outperforming MSSG~\cite{cheng2023language} by 21.3 points. 
Furthermore, We observe a similar trend on the \dLM-Grounding dataset, shown in Table~\ref{tab:drivelm_main}. 
Similar to \ttc, \ours shows significant improvements compared to directly finetuning from LLaVA-1.5, resulting in a 32.8 points improvement on the BEV AP$_\text{A}$ metric.

\subsection{Complex Reasoning in 3D}
\begin{table}[!t]
\centering
\caption{{\textbf{\dLM QA Benchmark}}. All models are finetuned on a subset of \dLM from held-out scenes. $\dagger$: we finetune LLaVA-1.5 on a subset of \dLM train split. 
\dLM baseline finetunes LLaMA Adapter V2~\cite{llama_adapter_v2} on a subset of \dLM train split. Top rows show models evaluated on the ``same split'' as the \dLM baseline. 
Bottom rows shows models evaluated on a larger test split (10 random scenes) held-out from all training. 
$\ddagger$: \dLM result on test data reported by authors. 
}
\resizebox{\textwidth}{!}{
\begin{tabular}{l>{\centering\arraybackslash}m{1.4cm}>{\centering\arraybackslash}m{1.4cm}>{\centering\arraybackslash}m{1.4cm}>{\centering\arraybackslash}m{1.4cm}>{\centering\arraybackslash}m{1.4cm}>{\centering\arraybackslash}m{1.4cm}>{\centering\arraybackslash}m{1.4cm}
}
\toprule
\multicolumn{1}{c}{Method}  & Acc. & ChatGPT & Match & BLEU$_1$ & ROUGE$_\text{L}$ & CIDEr & Overall \\
\midrule
\textit{baseline split} \\
\dLM baseline~\cite{sima2023drivelm} & 0.0 & {65.1} & 28.3 & 5.0 & 8.4 & 9.9 & 32.4 \\ 
LLaVA-1.5~\cite{liu2023improvedllava}$^\dagger$ & 38.5 & 53.5 & 26.1 & 15.8& 14.3 & 30.3 & 36.1 \\ 
\ours & \textbf{38.5} & \textbf{89.4} & \textbf{39.0} & \textbf{16.3} & \textbf{20.4} & \textbf{31.3} & \textbf{50.1} \\
\midrule 
\textit{our split} \\
\dLM baseline~\cite{sima2023drivelm}$^\ddagger$ & 0.0 & 67.8 & 18.8 & \textbf{23.8} & \textbf{19.9} & 0.7 & 32.8 \\ 
LLaVA-1.5~\cite{liu2023improvedllava}$^\dagger$ & 24.1 & \textbf{75.6} & 36.4 & 13.2 & 16.7 & 25.5 & 43.8 \\
\ours & \textbf{32.4} & 74.0 & \textbf{39.2} & {13.3}& 17.9 & \textbf{{25.6}} & \textbf{{45.4}} \\
\bottomrule
\end{tabular}
}
\vspace{-0.5cm}
\label{tab:drivelm_qa}
\end{table}
To show the effectiveness of 3D reasoning capability, we finetune \ours on \dLM-QA dataset. 
The dataset comprises questions about \textit{perception} (e.g., ``what are the objects worth noting in the current scenario?''), \textit{prediction} (e.g., ``where might the van, the sedan, and the pedestrian move in the future?), \textit{planning} (e.g., ``what are the safe actions of the ego car considering those objects?'') and \textit{behavior} (e.g., ``what would the ego vehicle's next action would be?''). 
We compare \ours with LLaVA-1.5~\cite{liu2023improvedllava} to show the impact of our pretraining, as well as the official baseline~\cite{sima2023drivelm} that has been recently released.
All models use 7-B scale LLM (Vicuna-7B~\cite{vicuna2023} or LLaMA-7B~\cite{touvron2023llama1}) and are fine-tuned on subset of \dLM train split. 
Top rows are the result on scenes held out by the authors and bottom rows are our additional split to evaluate models on a larger testset. 
The evaluation metric is based on accuracy, match (localization), BLEU/ROUGE$_\text{L}$/CIDEr, and ChatGPT score for favorable text generation. 
In both setting, \ours show competitive result consistently. 

\subsection{General MLLM Benchmarks}
We show the performance of \ours on general MLLM benchmarks. In Table~\ref{tab:rec}, we compare \ours to the state-of-the-arts in Referring Expression Comprehension (REC) benchmark on refCOCO/$\texttt{+}$/g~\cite{refcoco} dataset. 
We compare \ours to \textit{specialist} models such as MDETR~\cite{kamath2021mdetr} and UniTAB~\cite{yang2022unitab} which employs detection-specific architecture, and \textit{generalist} models of same size such as Ferret~\cite{you2023ferret}, Qwen-VL~\cite{Qwen-VL} and MiniGPT-v2~\cite{chen2023minigptv2}. 
In all test splits, \ours consistently outperforms with average score of \textbf{87.0}. 
In Table~\ref{tab:mllm}, we compare \ours with other competitive MLLMs of same model size on VQAv2~\cite{goyal2017making}, GQA~\cite{hudson2019gqa}, VizWiz~\cite{gurari2018vizwiz}, ScienceQA-Image~\cite{lu2022learn}, and POPE~\cite{Li-hallucination-2023}. 
The first row has models with fully zero-shot evaluation, and bottom rows has models that has seen images from some of the datasets. 
Compared to LLaVA-1.5~\cite{liu2023improvedllava}, miniGPT-v2~\cite{chen2023minigptv2} and Qwen-VL~\cite{Qwen-VL}, \ours maintain competitive result, validating that our 3D understanding does not degrade general reasoning capability of MLLM. 

\begin{figure}[!t]
    \centering
    \includegraphics[width=\linewidth]{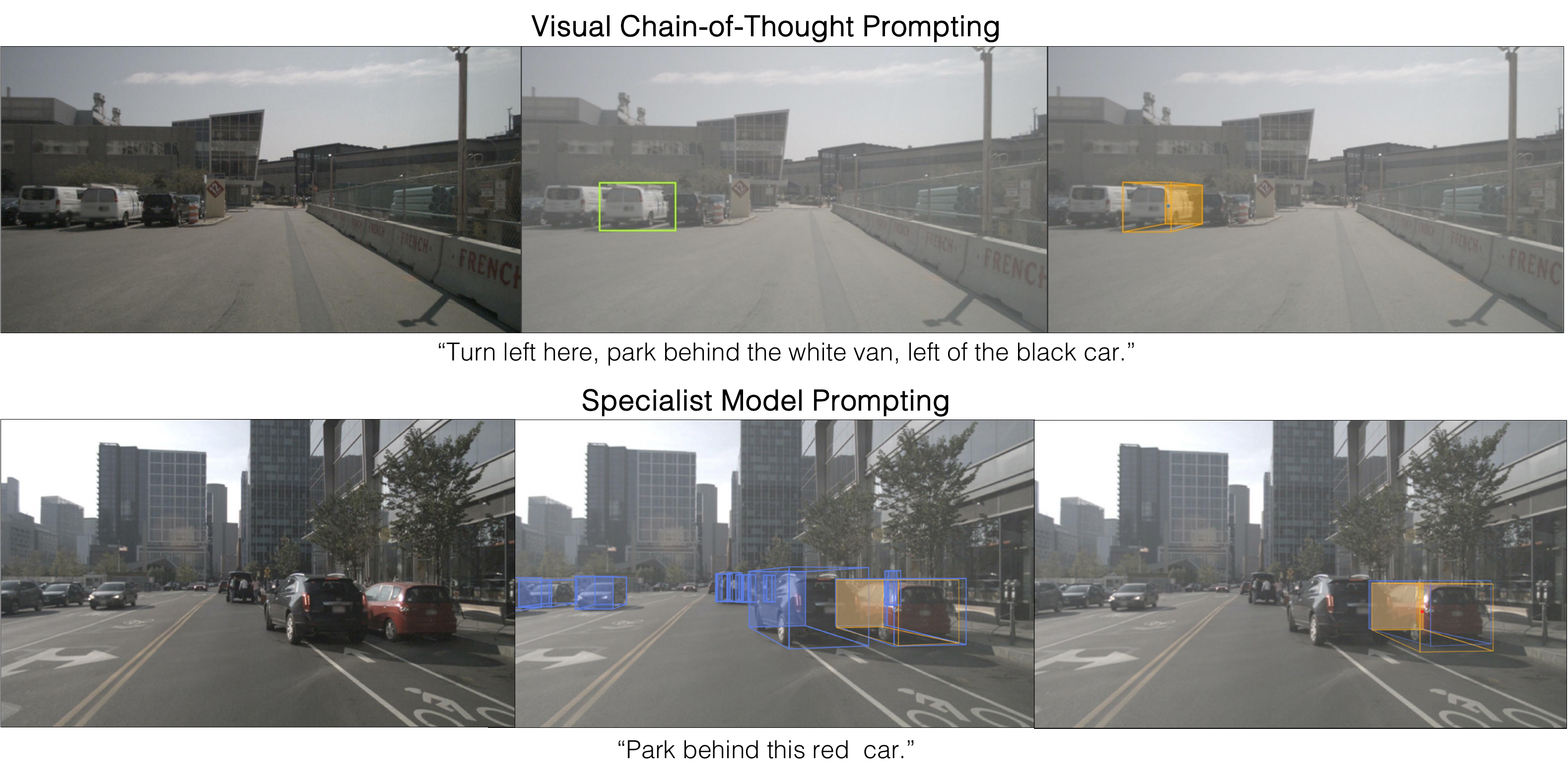}
    \vspace{-0.5cm}
    \caption{\textbf{\ours inference with prompting}. \textbf{Top}: Visual Chain-of-Thought Prompting to reason in 3D step-by-step. \textbf{Bottom}: Incorporating specialist models to further improve localization of \ours. Blue 3D boxes are the predictions of CenterPoint on corresponding LiDAR points. 
    }
    \label{fig:prompting_vis}
    \vspace{-0.8cm}
\end{figure}

\subsection{Ablation Study}
Our work consists of three key contributions, including a large-scale language-visual pre-training dataset~\dataset{}, visual chain-of-thought prompting, and specialist prompting. We provide comprehensive ablation studies for each of our design choices for \ours.

\vspace{0.2cm}\myparagraph{Data-scaling effect.} 
\dataset{} combines seven datasets, cumulatively encompassing 1.1 million images. To illustrate the impact of data-scaling on the performance, datasets are incrementally integrated during pre-training stage. We employ \ours without Visual Chain-of-Thought (VCoT) and fine-tune it at the image resolution $336 \times 336$. The results are presented in Table~\ref{tab:dataset_performance}. We observe a consistent improvement in performance, where AP grows from 19.7 to 44.7, improving by 25.0 points. 
It is also noteworthy that the incorporation of 2D data during pre-training benefits the model, even though the fine-tuning is performed for 3D grounding tasks - we observe a significant 6.3 points performance gain after including  refCOCO and COCO. 
Furthermore, in Table~\ref{tab:omni3d} we show the effect of increasing the pretraining data on indoor object grounding benchmarks. 
We evaluate \ours pre-trained on ``small'' subset of LV3D and the full LV3D, on Objectron~\cite{objectron2021}, ArkitScenes~\cite{dehghan2021arkitscenes}, and SUN-RGBD~\cite{song2015sun} datasets. 
The ``small'' LV3D and full LV3D contain the same amount of 3D indoor data (more details in the supplementary materials). 
We use the class name as text to prompt \ours, or class name and 3D location (\emph{left,center,right,close,far}). We describe the prompt generation process in more detail in our supplementary materials. 
We average the precision at IOU$_\text{3D}$ thresholds $\tau \in [0.15, 0.25, 0.5]$. 

\vspace{0.2cm}\myparagraph{Visual Chain-of-thoughts (VCoT) prompting during inference.} We evaluate \ours on \ttc with and without VCoT. As shown in Table~\ref{tab:vcot_ablation}, employing VCoT prompting yields significant enhancements in performance, with a improvement of (2.7, 1.9, 2.0, 2.3) points observed. 
It demonstrates that our VCoT is able to effectively bridge the gap between 2D semantic reasoning and 3D geometry reasoning compared to directly reasoning for 3D from text prompt. 
Figure~\ref{fig:prompting_vis} (top) visualizes the process. 

\vspace{0.2cm}\myparagraph{Impact of specialist prompting during inference.} 
Specialist prompting can leverage new input modality, such as LiDAR. 
As demonstrated in Table~\ref{tab:talk2car_main} on \ttc dataset, employing CenterPoint~\cite{yin2021center} predictions as visual prompts significantly improves the performance of \ours with gains of 25.1, 30.1, 29.4, 21.6 points in 3D grounding metrics. 
Note that \ours needs neither CenterPoint nor LiDAR points during training, which means user can choose any specialist model based on the available input modality. 
Figure~\ref{fig:prompting_vis} (bottom) visualizes the process.

\section{Conclusion}
In this paper, we present \ours, a multi-modal language model that can reason in both 2D and 3D. 
We provide a collection of dataset (\dataset{}) and a training framework to effectively scale \mmfm{} training for 3D understanding. 
We evaluate \ours in 2D and 3D grounded reasoning and VQA tasks, and show competitive results. 
We also show that \ours exhibits the behaviors of LLMs such as chain-of-thought prompting  to further improve 3D localization of our model. 
Finally, we show that our model can adapt any specialist models during inference by prompting their predictions as visual prompts. 
We examine that pure transformer-based \mmfm{} with minimal inductive bias can learn about 3D understanding solely by data scaling. 

\bibliographystyle{splncs04}
\bibliography{main}

\title{Language-Image Models with 3D Understanding - \emph{Supplementary Materials}} 

\author{Jang Hyun Cho\inst{1,2}\and
Boris Ivanovic\inst{2}\and
Yulong Cao\inst{2} \and
Edward Schmerling\inst{2} \and\\
Yue Wang\inst{2} \and 
Xinshuo Weng\inst{2} \and
Boyi Li\inst{2}\and
Yurong You\inst{2} \and\\
Philipp Krähenbühl\inst{1,\star} \and
Yan Wang\inst{2,\star} \and
Marco Pavone$^{2,}$\thanks{Equal advising}
}

\institute{UT Austin \and NVIDIA Research \\
\email{janghyuncho7@utexas.edu}\\
\email{\{bivanovic,yulongc,eschmerling,yuewang,xweng,\\
boyil,yurongy,yanwan,mpavone\}@nvidia.com}\\
}

\titlerunning{\ours}

\clearpage
\setcounter{page}{1}
\authorrunning{JH Cho et al.}

\maketitle

\section{Experiment Details}
In this section, we provide more details of our experiments.

\myparagraph{\dataset{}.} Each data in \dataset{} is an image and annotation pair.
Each annotation consists of a list of objects present in each image. 
Each object has a list of question and answer pairs as described in Section~\ref{sec:task_scale} of the main paper. 
If the data is from 2D dataset (\textit{e.g.}, COCO), the question answer pairs include ``$\texttt{text} \rightarrow \texttt{2D box}$'', ``$\texttt{2D center} \rightarrow \texttt{2D box}$'', ``$\texttt{2D box} \rightarrow \texttt{text}$'', \textit{etc}. 
Similarly, if the data is from 3D dataset (\textit{e.g.}, NuScenes), the question includes ``$\texttt{text} \rightarrow \texttt{3D box}$'', ``$\texttt{2D center} \rightarrow \texttt{3D box}$'', ``$\texttt{2D center} \rightarrow \texttt{depth}$'', ``$\texttt{2D box} \rightarrow \texttt{text}$'', \textit{etc}., as discussed in the Section~\ref{sec:method} of the main paper. 
To supplement text information, we leverage metadata from each dataset for each object class, such as object attribute in NuScenes dataset (``$\texttt{pedestrian}$'' $\rightarrow$ ``$\texttt{a walking pedestrian}.$''). 
For GRIT~\cite{peng2023kosmos}, we used the subset of the first 500 folders, which is about $\frac{1}{3}$. 
AS~\cite{wang2024allseeing_v2} is a collection of VQA datasets as well as some machine-generated 2D grounding data from a subset of SegmentAnything-1B~\cite{kirillov2023segment}. 
The original annotations contain a substantial amount of noise with duplicate answers. 
We simply remove the question-answer pairs of exactly identical and irrelevant answers. 
We also convert all the bounding boxes to follow the same format as \ours. 
For data standardization, we follow Omni3D~\cite{brazil2023omni3d} and convert all datasets to follow a virtual camera of focal length $f=512$. 

\myparagraph{\ours pre-training} undergoes pretraining stage and finetuning stage. 
The pretraining is done on \dataset{} with the dataset multiples specified in Table~\ref{tab:dataset_summary} of the main paper. 
In this stage, all object depth $z$ are transformed to align with the  virtual camera (same practice as Omni3D~\cite{brazil2023omni3d}) and converted to log-scale. 
For each ($x,y,z,w,h,l,r_1, r_2, r_3$), we normalize $x$ and $y$ in image coordinate from 0 to 999. 
For $z$, we set $z_\text{min}=-4$ and $z_\text{max}=5$ (after log-scale) and rescale in 0 and 999.
Similarly, $w_\text{min}=0, w_\text{max}=15, h_\text{min}=0, h_\text{max}=15, l_\text{min}=0, l_\text{max}=15$. 
All euler angles are normalized between $0$ and $2\pi$. 
We train all 3 Euler angles in ``yaw'', ``pitch'', and ``roll'' order.
Such ordering of angles in pretraining ensures the consistent sequential ordering before and after finetuning. 
To support flexible question formats during inference, we prepare a set of question templates and randomly sample one per object during training (\textit{e.g.}, ``\texttt{Provide 3D bounding box of the region in the image that this sentence describes: <>}'' or ``\texttt{What is the 3D box of the <>?}'').
For datasets where text does not contain orientation-specific information, we apply random horizontal flip data augmentation. 
We shuffle object order randomly, and use all objects even if there are duplicate questions, and cut off the training token sequence by the context length of the language model (4096). 
We pre-train with $336\times 336$ image size with frozen image-encoder, and $672\times672$ with full training. 

\myparagraph{\ours fine-tuning} undergoes a few change. Since finetuning benchmarks are all for outdoor scenes, we finetune $z$ to be in meter (\emph{i.e.,} no log-scale), and set $z_\text{min}=0, z_\text{max}=140$. We also ignore ``pitch'' and ``roll'' and only train for ``yaw'': ($x,y,z,w,h,l,r_1$). 
We finetune on \ttc, \dLM-grounding, and NuScenes dataset altogether for 10 epochs. 
We randomly prompt ground-truth boxes in the system prompt to allow specialist prompting at inference. 
We also randomly sample to query either 2D bounding box, 3D bounding box, or 2D-to-3D multi-turn question answering. 

\myparagraph{Indoor 3D grounding benchmark.} 
We use the testset of Objectron~\cite{objectron2021}, ArkitScenes~\cite{dehghan2021arkitscenes}, and SUN-RGBD~\cite{song2015sun} to evaluate the 3D grounding performance of \ours. In particular, we show the impact of data scaling with smaller subset of out pre-training dataset, LV3D-small. 
In LV3D-small, we remove the GRIT subset~\cite{peng2023kosmos}, AS-filtered~\cite{wang2024allseeing_v2}, Waymo~\cite{waymo}, Lyft~\cite{houston2021one}, Argoverse2~\cite{Argoverse2}, while both LV3D and LV3D-small have the same amount of indoor datasets. 
To evaluate grounding performance, we measure precision at $\tau$ where $\tau \in [0.15, 0.25, 0.5]$. 
When an image contains more than one objects associated to the input text prompt, we consider the max IOU. 
To augment object location to the text prompt, we add \texttt{``<object> close to camera''} if the depth is less than 0.8m. We add \texttt{``<object> on the left''} or \texttt{``<object> on the right''} if the object center is within the left/right 20 \% of the image and the distance from camera is 1/4/10 me away for small/medium/large objects. 
We define object as small/medium/large by the max dimension ($w,h,l$), with threshold of 0.5, 2, 3m. 
Similarly, we add \texttt{``<object> at the center''} if the object center is within the center 20 \% and the distance from camera is 1/4/10 m away for small/medium/large objects. 

\myparagraph{\dLM-QA training.} We aim to be consistent with the baseline training recipe~\cite{sima2023drivelm}. 
We preprocess \dLM questions and answers to follow the bounding box format of \ours; 3 decimal places, normalized between 0 and 1. 
For both LLaVA and \ours, we train on \dLM-QA for 5 epochs. 
For both LLaVA and \ours, we use image resolution of $336 \times 336$ and simply fed the 6 images independently to the vision encoder and concatenated before feeding to the language model. 
The number of vision tokens are $576 \times 6$ for each frame. 
We do not use any additional input (\textit{e.g.}, previous frames or point cloud) in order to compare to the baselines although \ours can enhance 3D perception with specialists. 
We hold out scene IDs: 

\small\texttt{"64a3a2d22172406c848f2a92275808ba"}, 
\small\texttt{"08be42eb2186411d8e2201225329f1c6"}, 

\small\texttt{"4b5bf3f4668d44fea9a676e9c4a8a79e"}, 
\small\texttt{"0e247ba64b9d4a34a7256b6c173b1b5d"},

\small\texttt{"dbd9183e1278475ea54761297e004b04"}, 
\small\texttt{"4098aaf3c7074e7d87285e2fc95369e0"},

\small\texttt{"9f3c8453d03d4df5946444757376b826"}, 
\small\texttt{"2fc3753772e241f2ab2cd16a784cc680"}, 

\small\texttt{"d0880a386b6d434bb5cd13c134af7a3e"}, 
\small\texttt{"01c3f5e39956402da3e37845632fadca"}

\noindent in \emph{our split} evaluation. 

\section{\ttc Grounding with VCoT.}
\begin{figure}[!t]
    \centering
    \includegraphics[width=\linewidth]{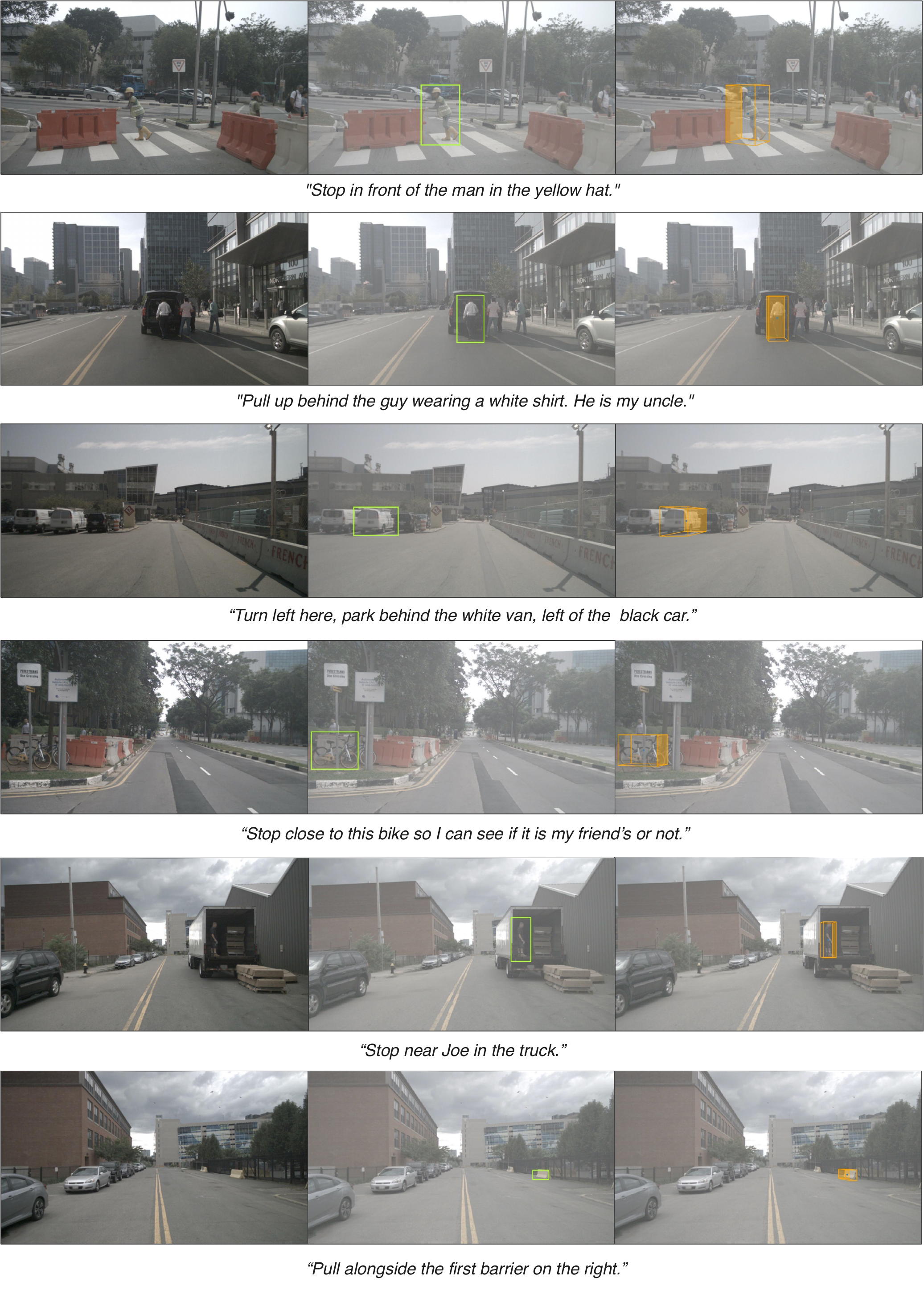}
    \caption{\textbf{\ours visual chain-of-thought prompting inference.} First column is input image, the second column is the 2D bounding box prediction, and the third column is the final 3D bounding box prediction prompted with the 2D prediction and text. } 
    \label{fig:vcot_vis}
\end{figure}
Figure~\ref{fig:vcot_vis} visualizes our visual chain-of-thought prompting inference on \ttc images. 
For each image and text prompt, we first ask with question: 

\texttt{``Please provide 2D bounding box of the region this sentence describes: <caption>.''}. 

\noindent Then, with the model prediction, we construct the second question as: 

\texttt{``Please provide 2D bounding box of the region this sentence describes: <caption>.''}

\texttt{<2D bounding box>}

\texttt{``Please provide 3D bounding box of the region this sentence describes: <caption>.''}
This simulates multi-turn conversation and the model can attend to the tokens of the previous conversation to infer the final output. 
We witness that as text prompt becomes more complicated, the guidance of 2D bounding box helps more.

\begin{figure}[!t]
    \centering
    \includegraphics[width=\linewidth]{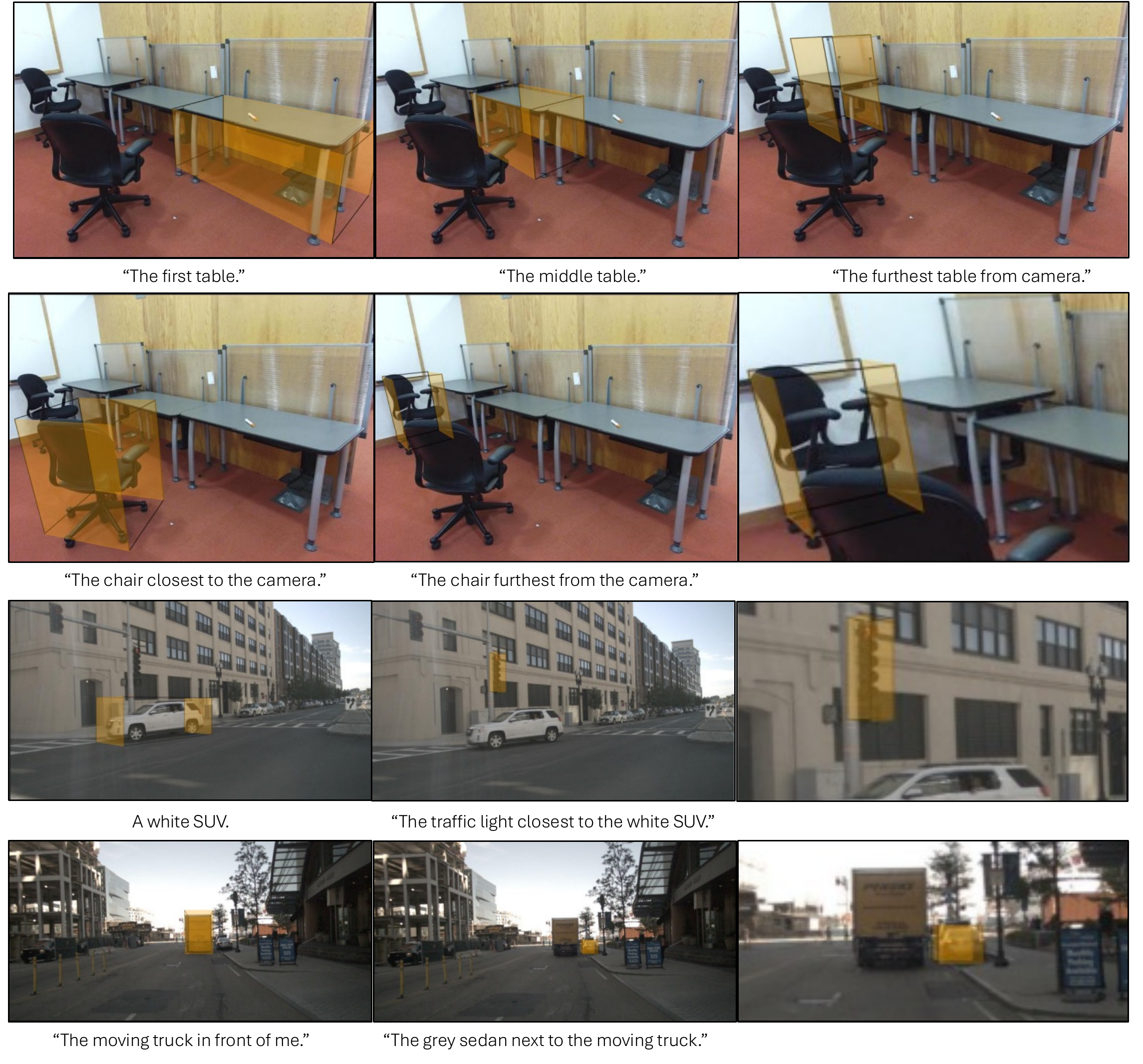}
    \includegraphics[width=\linewidth]{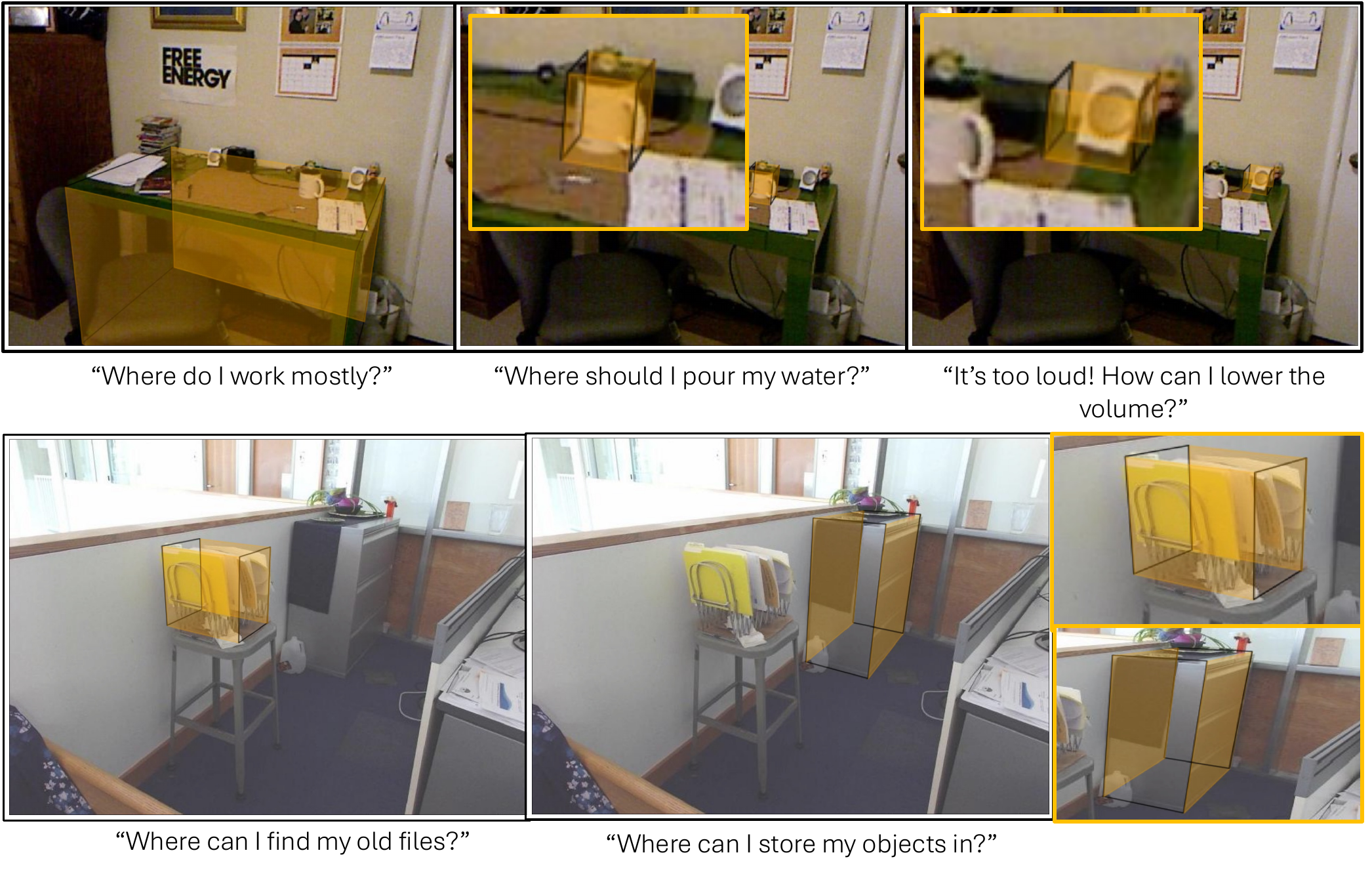}
    \caption{\textbf{More visualization of 3D grounding.} \ours is capable of grounding object with spatial cues and understand complex questions.}
    \label{fig:more_vis1}
\end{figure}

\begin{figure}[!t]
    \centering
    \includegraphics[width=\linewidth]{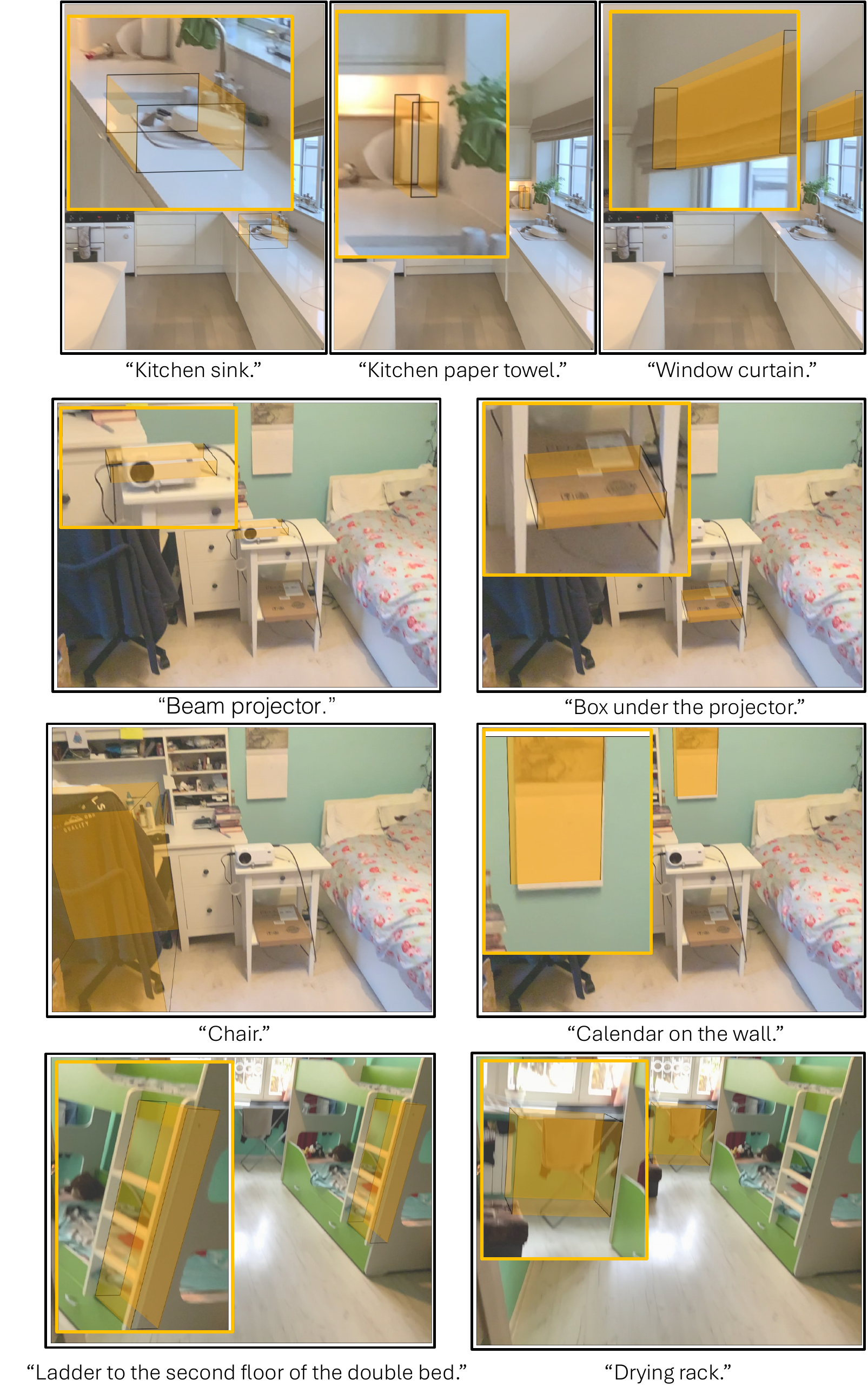}
    \caption{\textbf{More visualization of 3D grounding.} \ours is capable of grounding open-vocabulary category names.}
    \label{fig:more_vis2}
\end{figure}

\section{\dLM-QA Visualization}
\begin{figure}[!t]
    \centering
    \includegraphics[width=\linewidth]{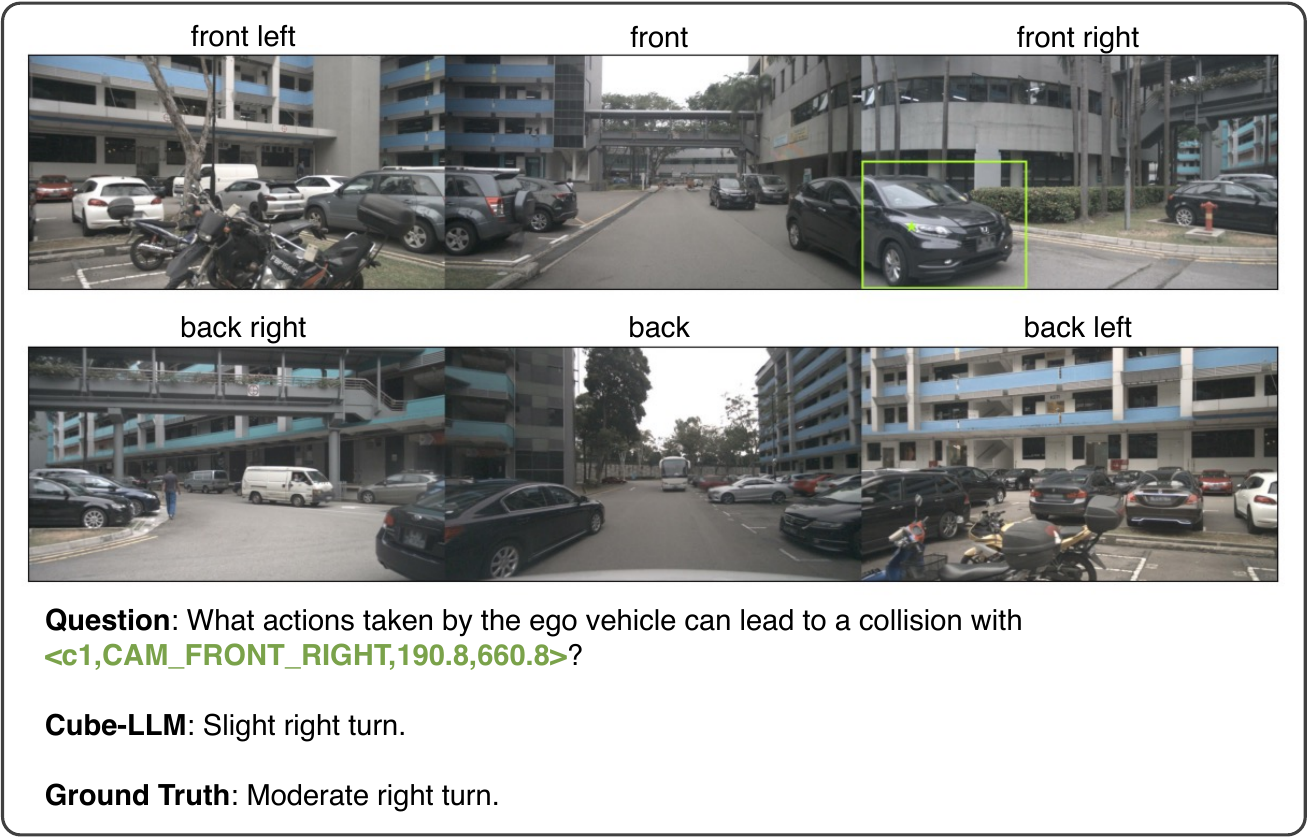}
    \includegraphics[width=\linewidth]{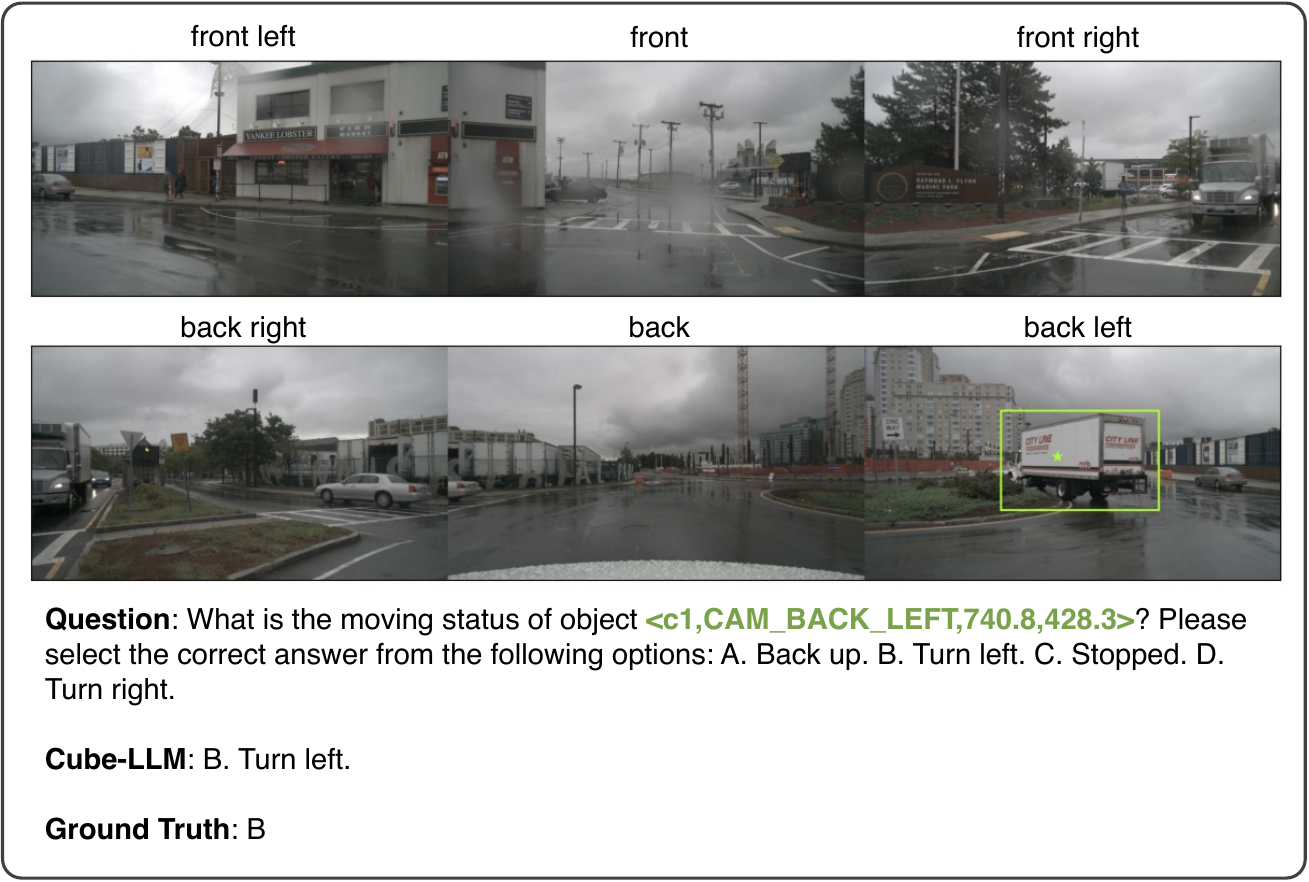}
    \caption{\textbf{\ours prediction on \dLM-QA.} \textcolor{mygreen}{Green marks} are the reference marks and the corresponding bounding box in the question. \textcolor{orange}{Orange marks} are predicted 2D points by \ours. 
    \textcolor{mytangoblue}{Blue marks} are the reference marks and the corresponding bounding box in the ground truth answers. }
    \label{fig:drivelm_qa_vis1}
\end{figure}

\begin{figure}[!t]
    \centering
    \includegraphics[width=\linewidth]{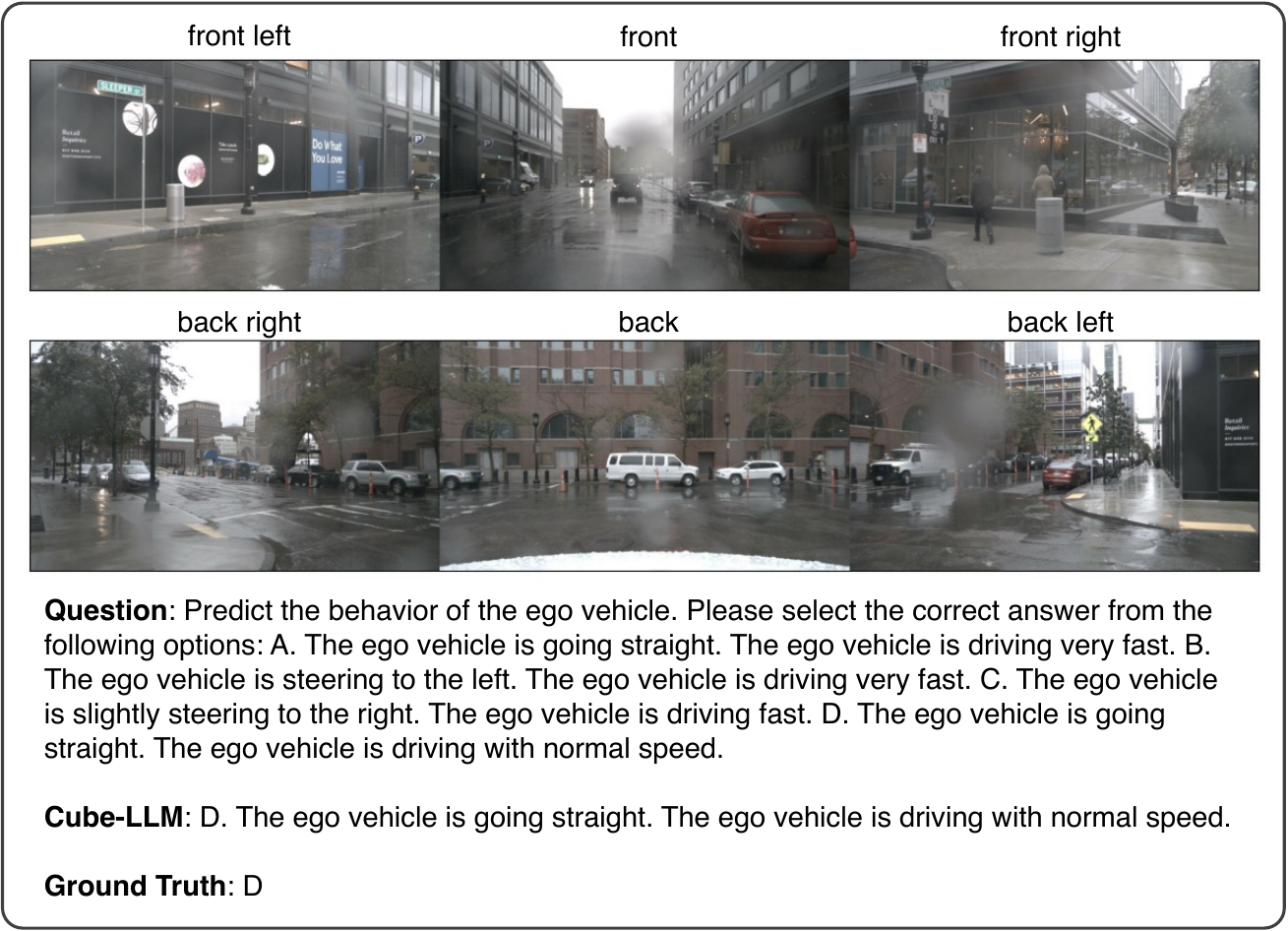}
    \includegraphics[width=\linewidth]{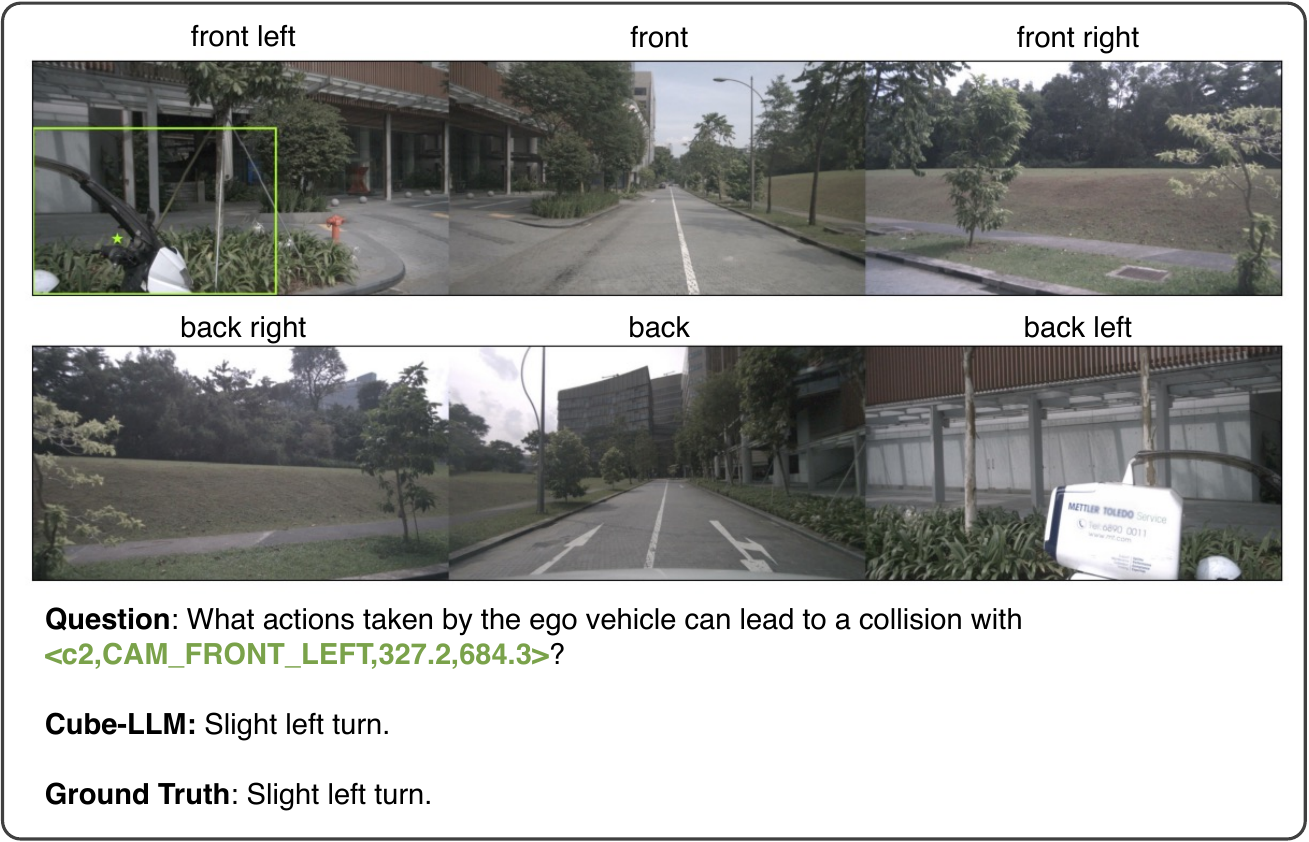}
    \caption{\textbf{\ours prediction on \dLM-QA.} \textcolor{mygreen}{Green marks} are the reference marks and the corresponding bounding box in the question. \textcolor{orange}{Orange marks} are predicted 2D points by \ours. 
    \textcolor{mytangoblue}{Blue marks} are the reference marks and the corresponding bounding box in the ground truth answers. }
    \label{fig:drivelm_qa_vis2}
\end{figure}

\begin{figure}[!t]
    \centering
    \includegraphics[width=\linewidth]{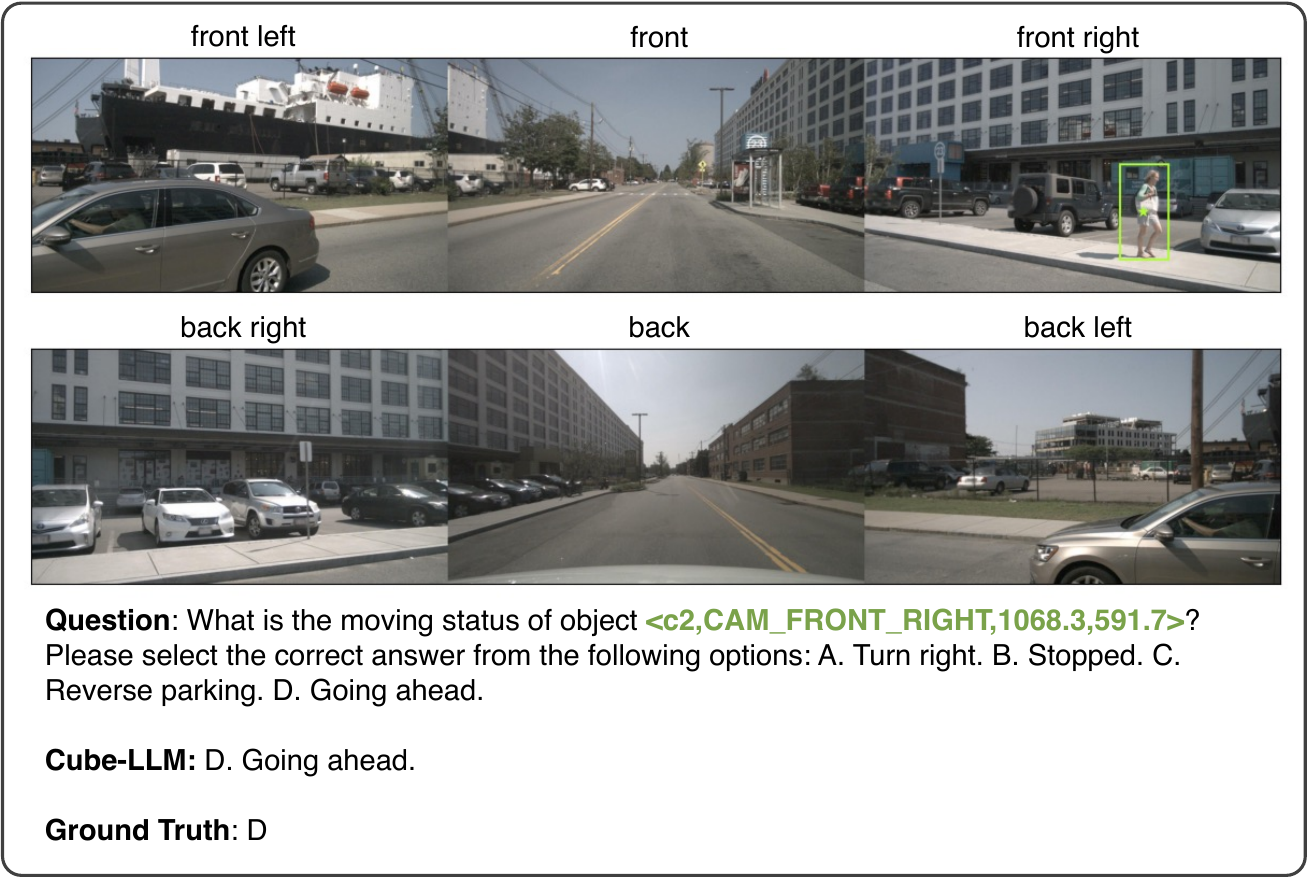}
    \includegraphics[width=\linewidth]{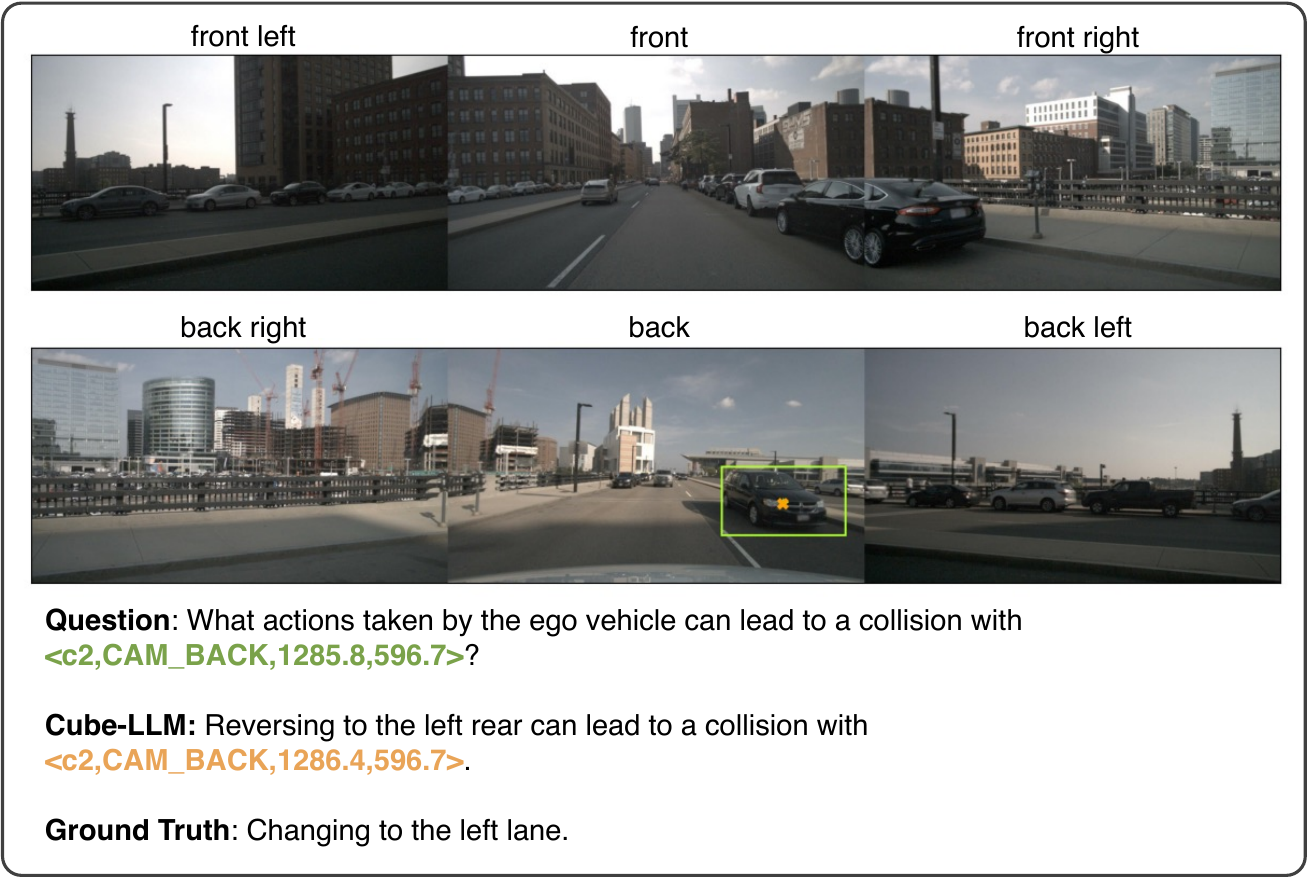}
    \caption{\textbf{\ours prediction on \dLM-QA.} \textcolor{mygreen}{Green marks} are the reference marks and the corresponding bounding box in the question. \textcolor{orange}{Orange marks} are predicted 2D points by \ours. 
    \textcolor{mytangoblue}{Blue marks} are the reference marks and the corresponding bounding box in the ground truth answers. }
    \label{fig:drivelm_qa_vis3}
\end{figure}

\begin{figure}[!t]
    \centering
    \includegraphics[width=\linewidth]{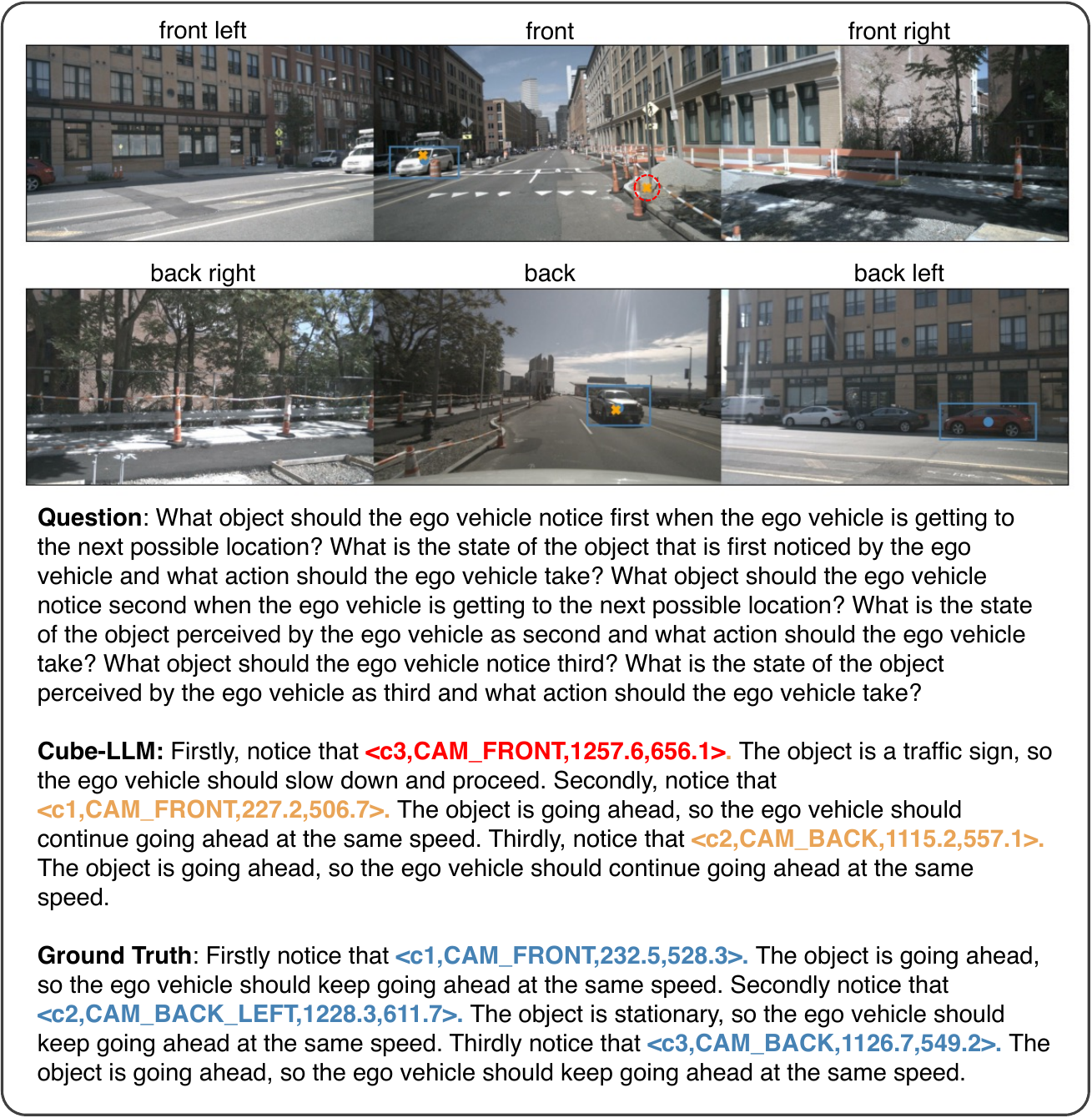}
    \caption{\textbf{\ours prediction on \dLM-QA.} \textcolor{mygreen}{Green marks} are the reference marks and the corresponding bounding box in the question. \textcolor{orange}{Orange marks} are predicted 2D points by \ours. 
    \textcolor{mytangoblue}{Blue marks} are the reference marks and the corresponding bounding box in the ground truth answers. \textcolor{red}{Red circle} is the predicted object that do not agree with the ground truth. }
    \label{fig:drivelm_qa_vis4}
\end{figure}

\begin{figure}[!t]
    \centering
    \includegraphics[width=\linewidth]{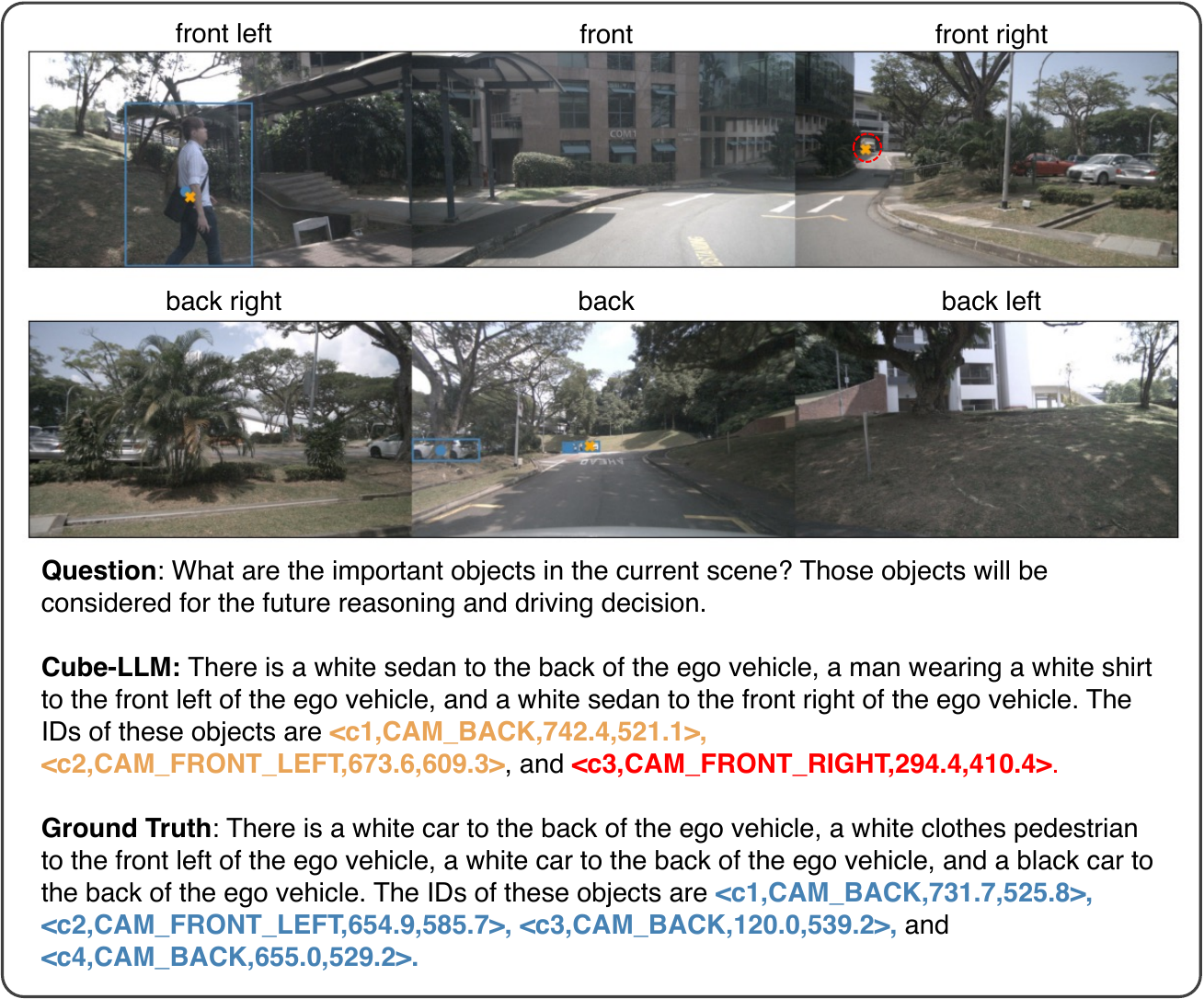}
    \caption{\textbf{\ours prediction on \dLM-QA.} \textcolor{mygreen}{Green marks} are the reference marks and the corresponding bounding box in the question. \textcolor{orange}{Orange marks} are predicted 2D points by \ours. 
    \textcolor{mytangoblue}{Blue marks} are the reference marks and the corresponding bounding box in the ground truth answers. \textcolor{red}{Red circle} is the predicted object that do not agree with the ground truth. }
    \label{fig:drivelm_qa_vis5}
\end{figure}

\begin{figure}[!t]
    \centering
    \includegraphics[width=\linewidth]{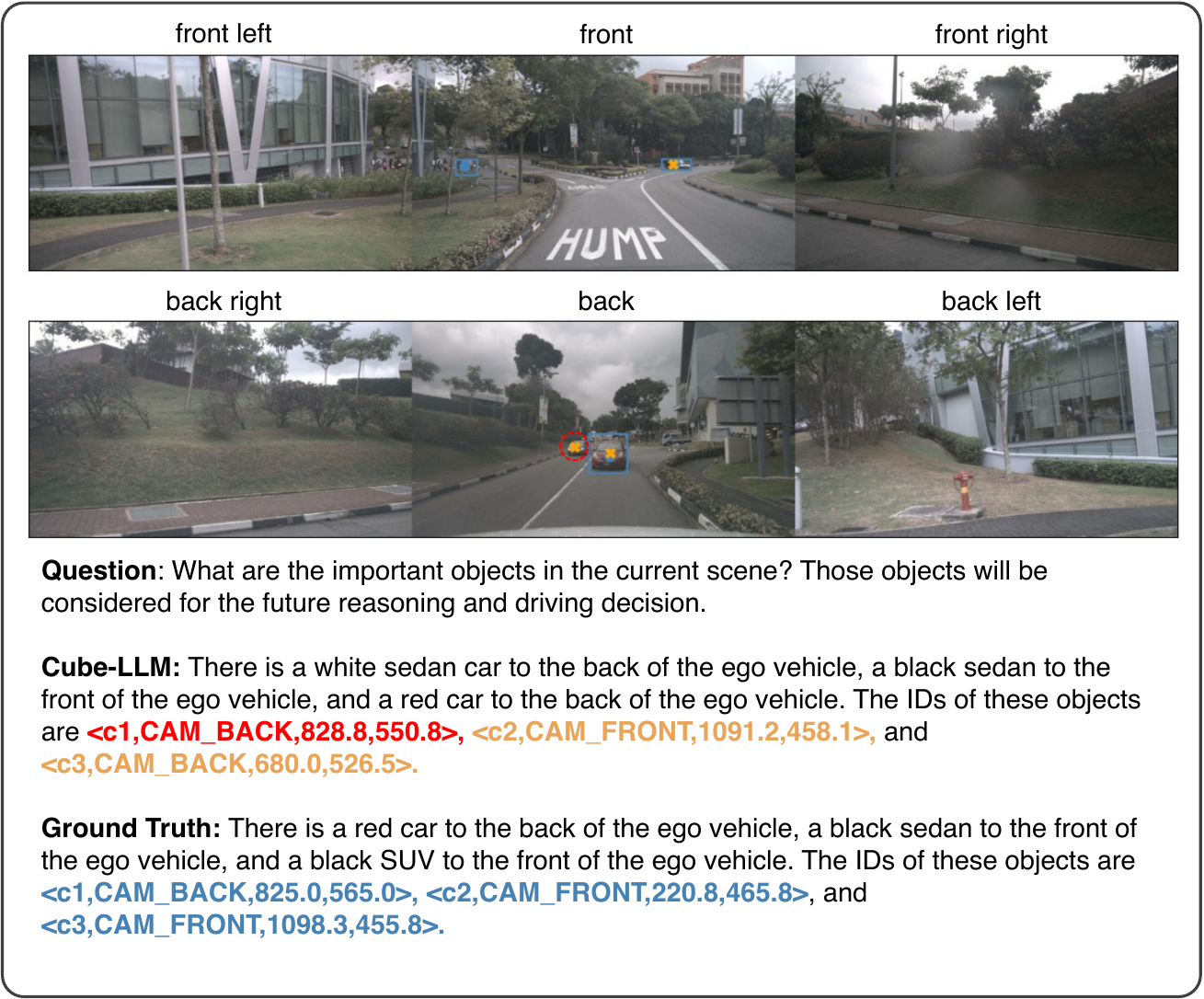}
    \caption{\textbf{\ours prediction on \dLM-QA.} \textcolor{mygreen}{Green marks} are the reference marks and the corresponding bounding box in the question. \textcolor{orange}{Orange marks} are predicted 2D points by \ours. 
    \textcolor{mytangoblue}{Blue marks} are the reference marks and the corresponding bounding box in the ground truth answers. \textcolor{red}{Red circle} is the predicted object that do not agree with the ground truth. }
    \label{fig:drivelm_qa_vis6}
\end{figure}

Figure~\ref{fig:drivelm_qa_vis1},~\ref{fig:drivelm_qa_vis2}, and~\ref{fig:drivelm_qa_vis3} show various types of \dLM questions. 
A large portion of the questions ask about a particular object specified in \texttt{<object ID, camera name, x, y>} format. 
\ours is capable of reasoning about the surrounding environment from the input multi-view images. 
When the \ours and the ground truth do not align (\textit{e.g.}, Figure~\ref{fig:drivelm_qa_vis1} top and~\ref{fig:drivelm_qa_vis3} bottom), it is evident that \ours understands the overall layout of surrounding objects relative to the ego vehicle.  
Figure~\ref{fig:drivelm_qa_vis4},~\ref{fig:drivelm_qa_vis5} and~\ref{fig:drivelm_qa_vis6} are the QA samples specifically for grounding important objects nearby. 
Notable points are that some of objects that \ours predicts that do not align with the ground truth (colored in \textcolor{red}{red}) are still important in each driving scenario. 
For example, in Figure~\ref{fig:drivelm_qa_vis4} \ours predicts a traffic sign (warning sign for cross road), in Figure~\ref{fig:drivelm_qa_vis5} \ours predicts a white sedan in front right camera that the ego may need to pay attention to, and in Figure~\ref{fig:drivelm_qa_vis6} \ours predicts a white sedan in back camera. 

\section{Failure Cases}

\begin{figure}[!t]
    \centering
    \includegraphics[width=\linewidth]{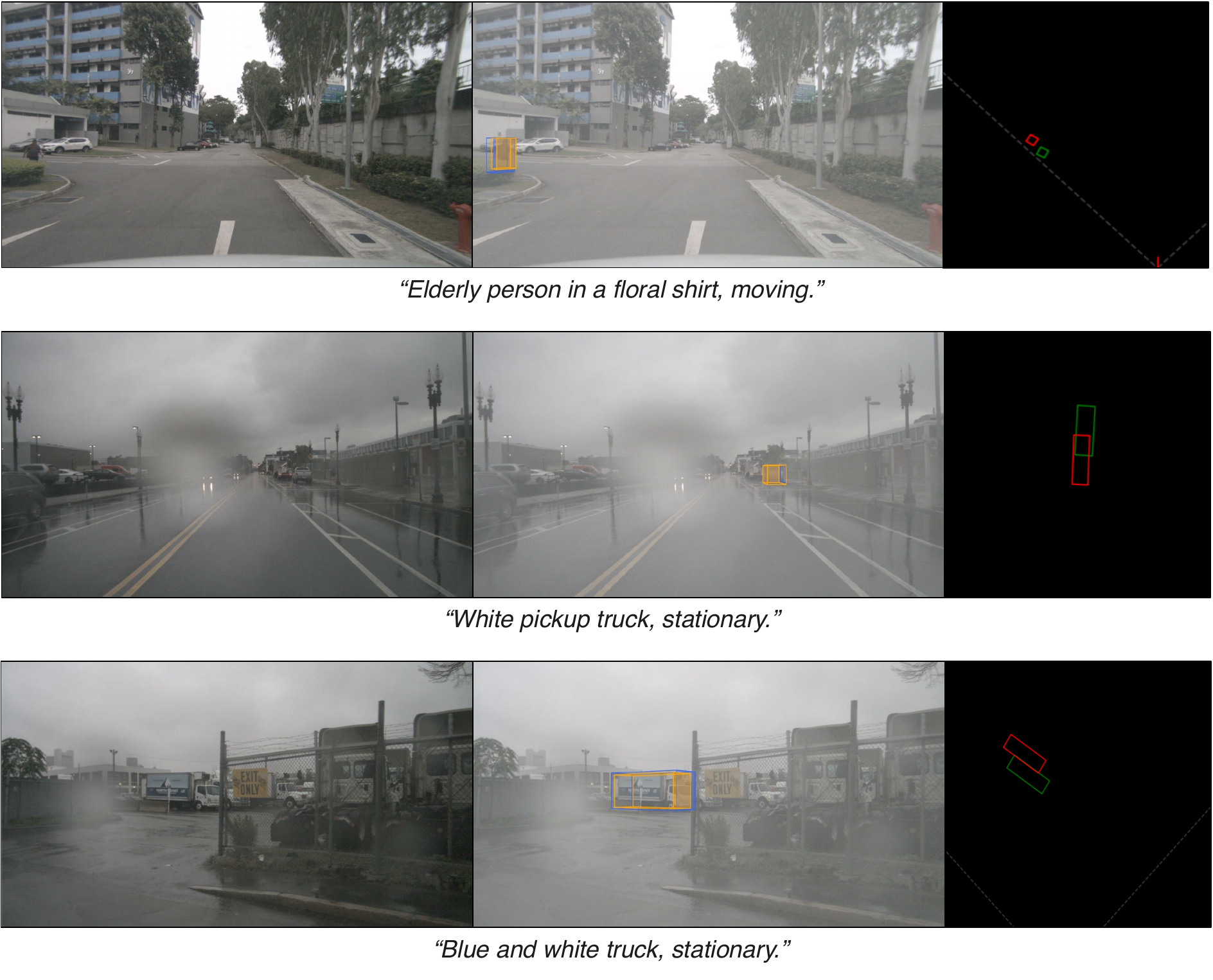}
    \caption{\textbf{Failure cases of \dLM-Grounding images.} The error mainly attributes to incorrect depth. Each row has the original image (left), projected 3D box prediction and ground truth (middle), and BEV image (right). 
    \textcolor{mytangoblue}{Blue box} is the ground truth and \textcolor{orange}{Orange box} is the prediction. 
    In BEV images, \textcolor{mygreen}{Green} box is the ground truth and \textcolor{myred}{red} box is the prediction. }
    \label{fig:drivelm_fail1}
\end{figure}

\begin{figure}[!t]
    \centering
    \includegraphics[width=\linewidth]{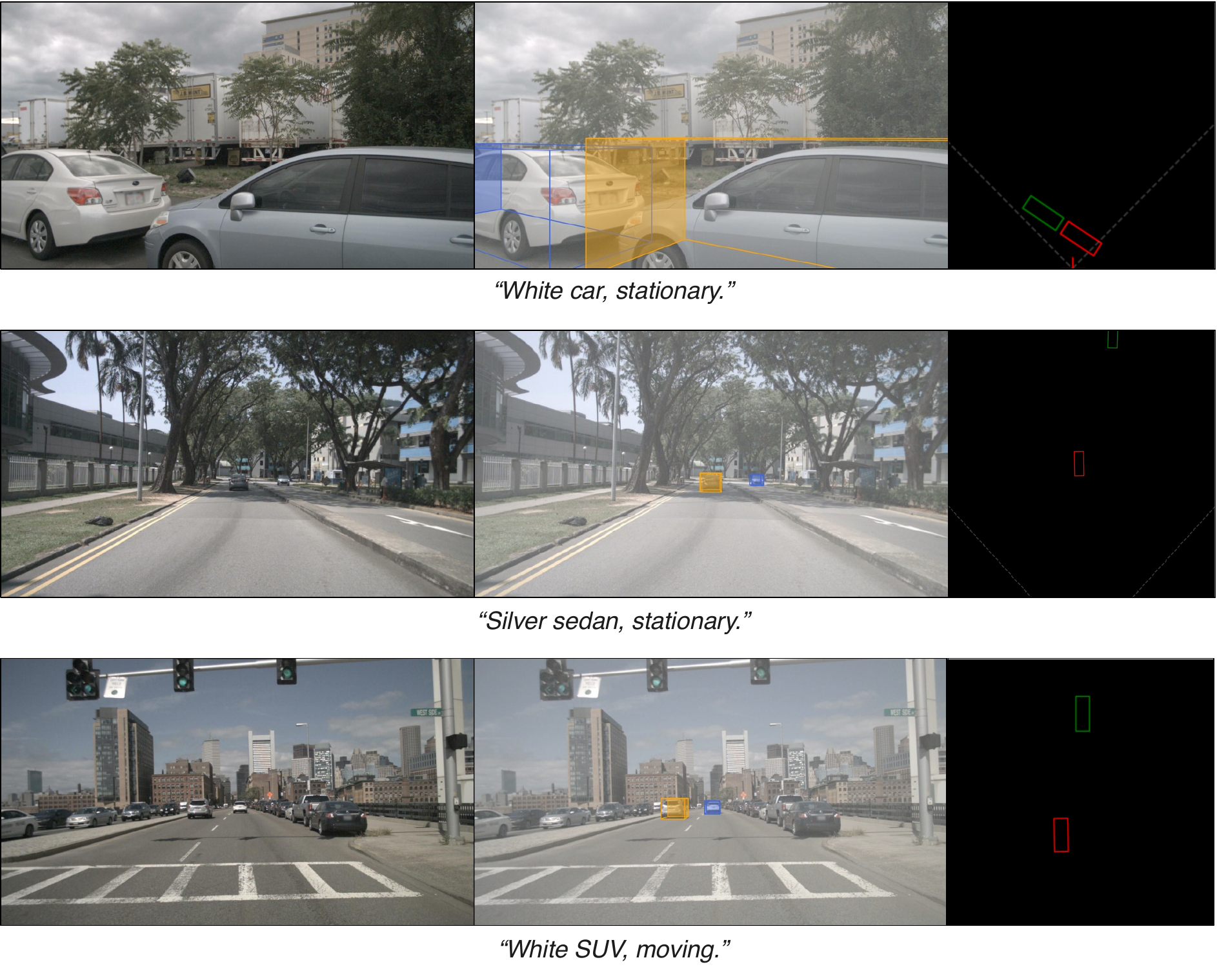}
     \caption{\textbf{Failure cases of \dLM-Grounding images.} The error mainly attributes to semantic mismatch. Each row has the original image (left), projected 3D box prediction and ground truth (middle), and BEV image (right). 
    \textcolor{mytangoblue}{Blue box} is the ground truth and \textcolor{orange}{Orange box} is the prediction. 
    In BEV images, \textcolor{mygreen}{Green} box is the ground truth and \textcolor{myred}{red} box is the prediction. }
    \label{fig:drivelm_fail2}
\end{figure}

\begin{figure}[!t]
    \centering
    \includegraphics[width=\linewidth]{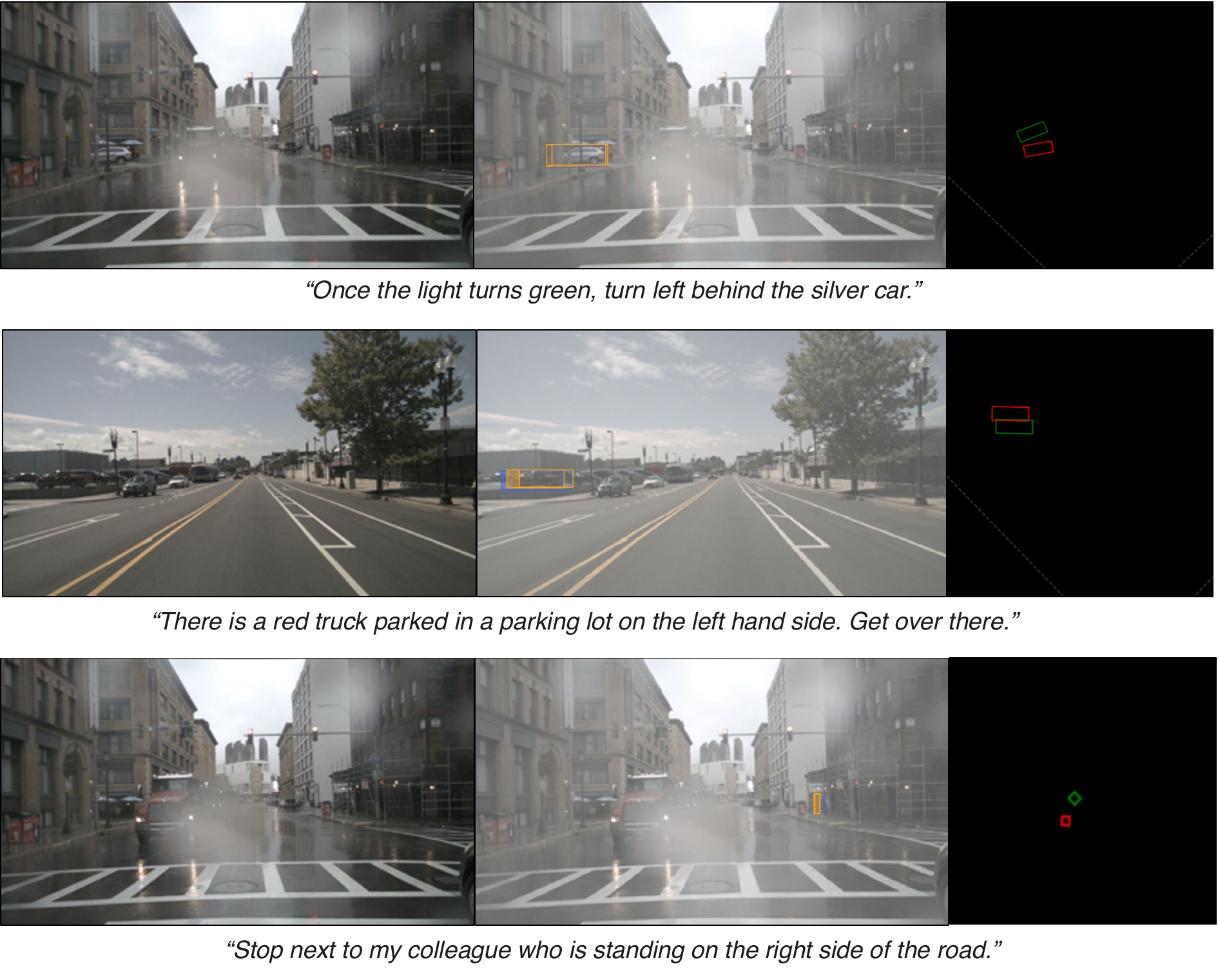}
    \caption{\textbf{Failure cases of \ttc images.} The error mainly attributes to incorrect depth. Each row has the original image (left), projected 3D box prediction and ground truth (middle), and BEV image (right). 
    \textcolor{mytangoblue}{Blue box} is the ground truth and \textcolor{orange}{Orange box} is the prediction. 
    In BEV images, \textcolor{mygreen}{Green} box is the ground truth and \textcolor{myred}{red} box is the prediction. }
    \label{fig:talk2car_fail2}
\end{figure}

\begin{figure}[!t]
    \centering
    \includegraphics[width=\linewidth]{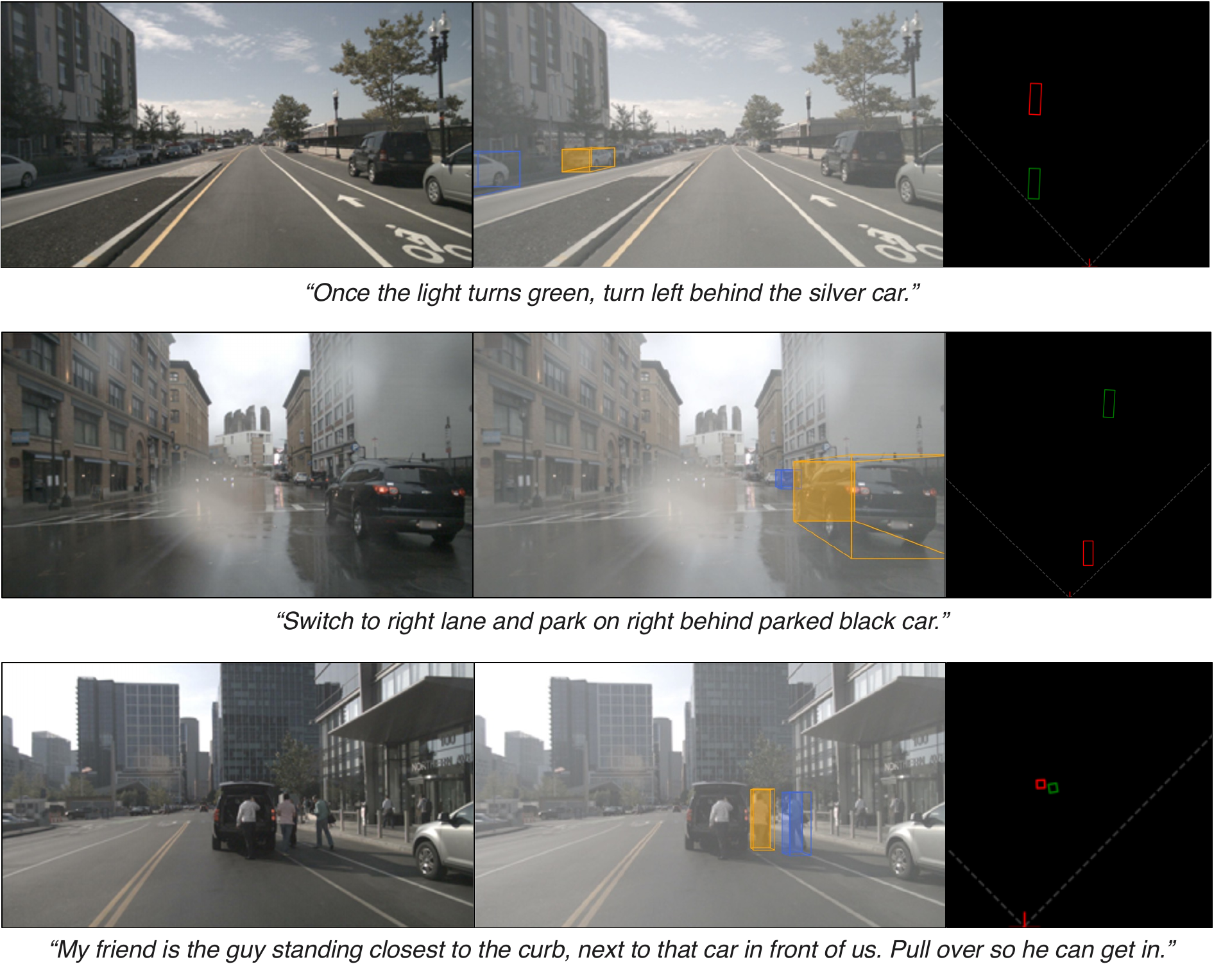}
    \caption{\textbf{Failure cases of \ttc images.} The error mainly attributes to semantic mismatch. Each row has the original image (left), projected 3D box prediction and ground truth (middle), and BEV image (right). 
    \textcolor{mytangoblue}{Blue box} is the ground truth and \textcolor{orange}{Orange box} is the prediction. 
    In BEV images, \textcolor{mygreen}{Green} box is the ground truth and \textcolor{myred}{red} box is the prediction.}
    \label{fig:talk2car_fail1}
\end{figure}

In Figure~\ref{fig:drivelm_fail1} and~\ref{fig:drivelm_fail2}, we show some failure cases of \ours grounding result on \dLM testset. 
\ours makes incorrect prediction mainly in two reasons: \emph{inaccurate depth} and \emph{semantic mismatch}. 
Figure~\ref{fig:drivelm_fail1} shows three examples of inaccurate depth errors and Figure~\ref{fig:drivelm_fail2} shows three examples of semantic mismatch. 
Notably, for the inaccurate depth cases, the projected 3D boxes show accurate 2D localization in image coordinate. 
This is because \ours trains to connect its 2D understanding to 3D, as described in Section~\ref{sec:v_cot} of the main paper. 
For the semantic mismatch cases, \ours struggles in correctly recognizing attributes when two similar objects are next to each other (\textit{e.g.}, \emph{silver sedan} vs. \emph{white sedan}, \emph{gray SUV} vs. \emph{white SUV}). 
Similarly, Figure~\ref{fig:talk2car_fail1} and Figure~\ref{fig:talk2car_fail2} show the failure cases of \ours on \ttc testset. 
Again, \ours still able to predict accurate size and projected 2D box region. 
Figure~\ref{fig:talk2car_fail2} show that \ours struggles recognizing the correct color of the car under the shade, the physical status of the black car (moving vs parked), and does not understand ``\emph{closest to the curb}.''

\section{Limitations}
\ours has several limitations. 
First, \ours does not employ any resampling methods~\cite{Dai2023InstructBLIPTG,Alayrac2022FlamingoAV} to reduce the number of vision tokens. 
This will limit the model to increase the input resolution to even larger than the current $672 \times 672$ (\textit{e.g.}, $1344 \times 1344$). 
\ours currently only supports a single frame input.
However, video input is critical in order to correctly recognize the dynamics of the environment. 
As a result, \ours tends to fail correctly predicting whether an object is stationary or moving, or rely on the location of object in the scene and infer the object's dynamics (\textit{e.g.}, a car inside a parking space is most likely stationary). 
We leave these limitations for the future work. 

\end{document}